\documentclass[lettersize,journal]{IEEEtran}
\usepackage[top=1in, bottom=1.04in, left=1.01in, right=1.01in]{geometry}
\usepackage{setspace}
\usepackage{amsthm}

\usepackage{fancyhdr}

\newcommand{\bfb}{\mathbf{b}}

\newcommand{\bfc}{\mathbf{c}}
\newcommand{\bfs}{\mathbf{s}}
\newcommand{\bfx}{\mathbf{x}}

\newcommand{\bfp}{\mathbf{p}}

\newcommand{\calB}{\mathcal{B}}
\newcommand{\calD}{\mathcal{D}}
\newcommand{\calM}{\mathcal{M}}

\newcommand{\calE}{\mathcal{E}}

\newcommand{\calQ}{\mathcal{Q}}

\newcommand{\calO}{\mathcal{O}}
\newcommand{\bbR}{\mathbb{R}}
\newcommand{\bbC}{\mathbb{C}}
\newcommand{\bbN}{\mathbb{N}}
\newcommand{\bbZ}{\mathbb{Z}}
\newcommand{\bsth}{\boldsymbol{\theta}}
\newcommand{\bsa}{\boldsymbol{\alpha}}
\newcommand{\bsb}{\boldsymbol{\beta}}
\newcommand{\bfa}{\mathbf{a}}
\newcommand{\bbE}{\mathbb{E}}

\newcommand{\netUE}{g}
\newcommand{\netAP}{h}
\newcommand{\netWH}{f}
\newcommand{\parUE}{\boldsymbol{\gamma}}
\newcommand{\parAP}{\boldsymbol{\phi}}
\newcommand{\ic}{q}

\newtheorem{remark}{\textbf{Remark}}

\newtheorem{lemma}{\textbf{Lemma}}
\newtheorem{theorem}{\textbf{Theorem}}

\ifCLASSINFOpdf
  \usepackage[pdftex]{graphicx}
\else
  \usepackage[dvips]{graphicx}
\fi
\usepackage{amsmath}
\usepackage{amssymb}

\allowdisplaybreaks

\usepackage{algorithmic}
\usepackage{algorithm}
\ifCLASSOPTIONcompsoc
 \usepackage[caption=false,font=normalsize,labelfont=sf,textfont=sf]{subfig}
\else
 \usepackage[caption=false,font=footnotesize]{subfig}
\fi
\usepackage{color,cite}

\newcounter{relctr} 

\newcommand\labelrel[2]{%
  \begingroup
    \refstepcounter{relctr}%
    ~\underset{\textnormal{(\alph{relctr})}}{\mathstrut{#1}}~%
    \originallabel{#2}%
  \endgroup
}
\AtBeginDocument{\let\originallabel\label} 

\allowdisplaybreaks

\fancypagestyle{firstpage}
{
    \fancyhf{}
    \fancyhead[C]{\footnotesize This work has been submitted to the IEEE for possible publication. Copyright may be transferred without notice, after which this version may no longer be accessible.}
    
}

\begin{document}
\bstctlcite{IEEEexample:BSTcontrol}

%
\title{Pigeon-SL: Robust Split Learning Framework for Edge Intelligence under Malicious Clients}
%
%
%

\author{Sangjun Park,~\IEEEmembership{Member,~IEEE}, Tony Q.S. Quek,~\IEEEmembership{Fellow,~IEEE}, and Hyowoon Seo,~\IEEEmembership{Member,~IEEE}
        \vspace{-20pt}

		\thanks{
			S.~Park is with the Department of Electronics and Communications Engineering, Kwangwoon University, Seoul 01897, Korea (e-mail: sangjunpark@kw.ac.kr). 
		}
        \thanks{
        Tony Q. S. Quek is with the Singapore University of Technology and Design, Singapore 487372 (e-mail: tonyquek@sutd.edu.sg).
        }
        \thanks{H.~Seo is with the Department of Electrical and Computer Engineering, Sungkyunkwan University, Suwon 16419, South Korea (e-mail: hyowoonseo@skku.edu)}
        \thanks{(\emph{Corresponding Author: Hyowoon Seo})}
}

\maketitle
\thispagestyle{firstpage}
\begin{abstract}
Recent advances in split learning (SL) have established it as a promising framework for privacy-preserving, communication-efficient distributed learning at the network edge. However, SL’s sequential update process is vulnerable to even a single malicious client, which can significantly degrade model accuracy. To address this, we introduce \emph{Pigeon-SL}, a novel scheme grounded in the pigeonhole principle that guarantees at least one entirely honest cluster among $M$ clients, even when up to $N$ of them are adversarial. In each global round, the access point partitions the clients into $N+1$ clusters, trains each cluster independently via vanilla SL, and evaluates their validation losses on a shared dataset. Only the cluster with the lowest loss advances, thereby isolating and discarding malicious updates. We further enhance training and communication efficiency with \emph{Pigeon-SL+}, which repeats training on the selected cluster to match the update throughput of standard SL. We validate the robustness and effectiveness of our approach under three representative attack models-label flipping, activation and gradient manipulation—demonstrating significant improvements in accuracy and resilience over baseline SL methods in future intelligent wireless networks.
        \vspace{-10pt}
\end{abstract}


%
\IEEEpeerreviewmaketitle

\section{Introduction}

Distributed machine learning has become essential for processing large-scale datasets and compute-intensive tasks across diverse domains such as healthcare, finance, and robotics, while preserving data privacy and reducing communication overhead. Traditional centralized training, which aggregates raw data at a central server, is often impractical due to privacy concerns, bandwidth limitations, and the volume of edge-generated data. Federated learning (FL) \cite{arXiv15_Konecny,SIGSAC15_Shokri,AISTATS17_Mcmahan} addresses some of these challenges by exchanging model updates instead of raw data, though it still incurs significant communication costs when transmitting full gradients or weights.

To further alleviate these limitations, split learning (SL) \cite{SL_conv,SL_conv02,ML_lab_02} has emerged as a complementary paradigm that partitions the model into client-side and server-side components. SL transmits only activations and cut-layer gradients, keeping raw data entirely local. This design dramatically reduces communication load and is especially well-suited for resource-constrained, privacy-sensitive environments.

Building on these paradigms, SplitFed Learning (SFL) \cite{splitfed} combines the strengths of FL and SL. It introduces a hybrid architecture where clients train up to the cut layer (as in SL), and a federated server aggregates client-side models via federated averaging. This setup can improve generalization under certain system configurations. However, like FL, SFL exposes client-side models to the server, which may pose privacy risks in sensitive applications. In contrast, SL’s unique advantage lies in sharing only intermediate representations, underscoring its potential for secure learning and motivating further investigation.

Despite these advantages, SL presents its own challenges. Its inherently sequential update process means that a single malicious client can inject corrupted activations or gradients, severely disrupting convergence and degrading model performance. While FL has inspired a wide range of Byzantine-robust aggregation techniques and reputation-based defenses \cite{FL_att_01,FL_att_02,FL_att_03,FL_att_04,FL_att_05,FL_att_06,FL_att_07}, analogous defense mechanisms for SL remain largely underexplored. This gap highlights the urgent need to develop robust SL frameworks that can withstand adversarial behaviors.

To address the lack of robustness in SL, we propose \textit{Pigeon-SL}, a novel cluster-based defense that leverages the pigeonhole principle to isolate honest updates even when up to $N$ out of $M$ clients are adversarial. In each global round, the access point (AP) randomly partitions the $M$ clients into $N{+}1$ clusters, independently trains each via vanilla SL, and evaluates their validation losses on a shared dataset. Since all clusters are assessed on the same dataset, the validation loss provides a fair metric of training quality. By selecting the cluster with the lowest loss, Pigeon-SL discards malicious contributions while ensuring at least one entirely honest cluster remains.

To compensate for the reduced per-round update throughput caused by clustering, we introduce \textit{Pigeon-SL+}. After selecting the best-performing cluster, Pigeon-SL+ continues SL training on it for $N$ additional subrounds, yielding $N{+}1$ updates per global iteration—matching the throughput of standard SL while preserving robustness.

We provide a convergence analysis of Pigeon-SL under standard smoothness and bounded variance assumptions, showing sublinear regret and convergence to a stationary point at rates comparable to existing distributed learning methods. Through simulations of three representative attacks—label flipping, activation tampering, and gradient tampering—we demonstrate that our methods significantly outperform baseline SL in both accuracy and resilience.

\subsection{Related Work}
\subsubsection{Split Learning (SL)}
SL was first introduced in \cite{SL_conv,SL_conv02} and demonstrated superior communication and computation efficiency compared to other distributed learning paradigms. Since then, SL has been extended in various directions to tackle emerging challenges and applications \cite{SL01}. For example, \cite{splitfed} combines SL with FL to accelerate training, while \cite{SL04} introduces communication-aware SL to address channel variability and device heterogeneity in IoT platforms. Parallel SL frameworks with integrated resource management have also been proposed to reduce training latency in wireless edge environments \cite{SL_resource, SL03}. To enhance speed and privacy for mobile applications, \cite{SL_Binarizing} proposes binarizing client-side layers. In terms of privacy, pruning-based SL with differential privacy is introduced in \cite{SL02} for secure learning in satellite networks. Moreover, vulnerabilities arising from a malicious server have been identified \cite{SL_privacy_01,SL_privacy_02,SL_privacy_03}, and corresponding defense mechanisms have been developed.

Despite these advances, comparatively little attention has been given to defending SL systems against attacks from malicious clients—a critical gap our work aims to address.

\subsubsection{Robust Distributed Learning}
To address the challenges posed by malicious clients, extensive research has focused on enhancing robustness in distributed learning—particularly within the FL framework. For instance, \cite{FL_att_04} proposes a hierarchical audit-based FL scheme for improved reliability and security with minimal overhead, while \cite{FL_att_05} introduces reputation-based aggregation and resource optimization to boost training efficiency in wireless FL. In \cite{FL_att_07}, a proactive defense mechanism is developed using a federated analytics paradigm. Additional defense strategies against malicious attacks in FL are presented in \cite{FL_att_01, FL_att_02, FL_att_03, FL_att_06}.

Although robustness has been actively studied in FL, similar efforts for SL remain limited. To safely harness the efficiency advantages of SL in distributed systems, it is crucial to develop robust SL techniques that can withstand adversarial clients.

\subsection{Contributions, Organization and Notations}
\subsubsection{Summarized Contributions}
\begin{itemize}
    \item We propose a robust client clustering approach that identifies trustworthy client groups by monitoring cluster-level validation losses. Model updates are performed using the most reliable cluster to mitigate the influence of malicious clients.

    \item To safeguard the validation process, we introduce a tamper-resilient mechanism using shared validation samples, preventing adversaries from manipulating loss values or cluster selection.

    \item We validate the effectiveness of our scheme under three representative tampering attacks, demonstrating superior resilience compared to conventional methods.
\end{itemize}
\subsubsection{Organization} The remainder of the paper is organized as follows. Section II describes the SL system model. Section III presents the proposed Pigeon-SL method. Section IV provides its convergence analysis. Section V reports simulation results, and Section VI concludes the paper.

\subsubsection{Notations}
Throughout the article, scalars are written in a normal font, and vectors are written in a bold font. $\bbN$, $\bbZ^+$, $\bbR$ and $\bbC$ are sets of natural, positive integer, real and complex numbers, respectively. The $L^2$-norm is denoted as $\lVert\cdot\rVert$ for vectors and $\lvert\cdot\rvert$ for scalars, while $(\cdot)^{\mathsf{T}}$ represents the transpose. For $a\in\bbN$, $[a]$  denotes $\{n \mid n\in\bbN, 1\le n\le a\}$. For $\bfx\in\bbR^{n}$ and the scalar-valued function $f(\bfx)\in\bbR$, $\nabla_{\!\bfx} f(\bfx)\in\bbR^{n}$ denotes the gradient of $f(\bfx)$ with respect to $\bfx$.

\begin{figure}[t]
    \centering
    \subfloat{\includegraphics[width=0.48\textwidth]{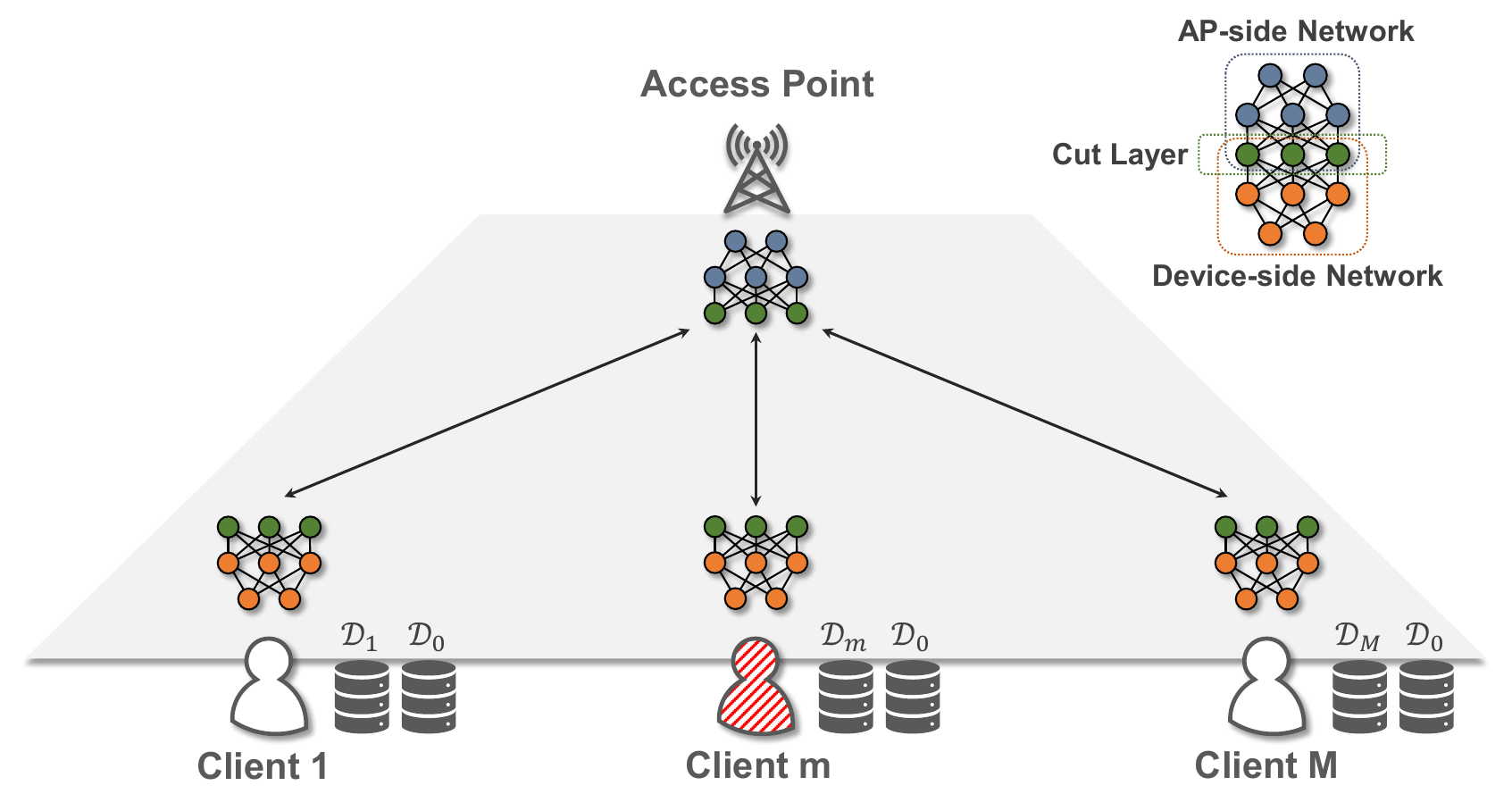}\label{fig:syst}} 
    \caption{An illustration of a SL network with the malicious clients.}
    \label{fig:systemmodel}
\end{figure}

\section{System Model}
Consider an SL scenario over a network consisting of a single AP and $M\in\mathbb{Z}^+$ distributed clients, as shown in Fig.~\ref{fig:systemmodel}. Each client $m\in[M]$ holds a local dataset $\mathcal{D}_m =\bigl\{\mathbf{s}_m^i=(\mathbf{x}_m^i,y_m^i)\mid i \in [D_m]\bigr\}$,
where $\mathbf{x}_m^i$ and $y_m^i$ denote the input and corresponding label, respectively, drawn i.i.d. from the distribution $p(\mathbf{x},y)$. We denote the total number of local samples by $D =\sum_{m=1}^M D_m$.

In addition, all clients share a common (reference) dataset $\mathcal{D}_o =\bigl\{\mathbf{s}_o^i=(\mathbf{x}_o^i,y_o^i)\mid i \in [D_o] \bigr\}$, with $D_o$ samples. {To construct $\mathcal{D}_o$, the AP samples data from $p(\mathbf{x}, y)$ and broadcasts it to all clients prior to training.} Subsequently, $\mathcal{D}_o$ serves as a critical mechanism to safeguard the learning process against performance degradation caused by malicious participants.



The AP and clients collaborate to train a model for a given downstream task. Specifically, a local loss function of the client $m$ is denoted as $L_m(\bsth)=\frac{1}{D_m}\sum_{\bfs\in\calD_m} \ell(\bfs;\bsth)$ where $\ell(\cdot;\bsth)$ is a sample-wise loss function\footnote{In our experiments, we adopt cross-entropy as the sample-wise loss function, as the primary focus is on classification tasks.} parameterized by $\bsth\in\bbR^d$. Hence, the goal of this distributed learning network is to determine the optimal parameters $\bsth^*\in\bbR^d$ that minimize the overall loss function, defined by $L(\bsth) = \sum_{m\in\calM}\frac{D_m}{D}L_m(\bsth)$.

The learning process follows a sequential approach across the clients similar to vanilla SL \cite{SL_conv}.
A global round is defined as a single learning iteration performed by all clients, while a mini-batch update is defined as a the learning process within each individual client on a single mini-batch of data. In each global round, once a client completes its local update, the learning process moves to the next client's turn in a sequential manner. For simplicity, we assume that our proposed SL runs for $T\in\bbZ^+$ global rounds and $E\in\bbZ^+$ mini-batch updates. 

The training neural network (NN) model is split into two segments at a cut layer: one belonging to the clients and the other to the AP. The cut layer is of size $d_{\bfc}\in\bbZ^+$, with the output of the client-side NN acting as the input to the AP-side NN. The client-side and AP-side NNs contain $d_{\text{CL}}\in\bbZ^+$ and $d_{\text{AP}}\in\bbZ^+$ parameters, respectively. Specifically, the overall parameters $\bsth$ are divided across the cut layer into two disjoint sets: $\parUE\in\bbR^{d_{\text{CL}}}$ for the client-side NN preceding the cut layer and $\parAP\in\bbR^{d_{\text{AP}}}$ for the AP-side NN succeeding it. The output of the client-side NN, given the input data $\bfx_{\text{CL}}$ is defined as $g(\bfx_{\text{CL}},\parUE)\in\bbR^{d_{\bfc}}$. Similarly, the output of the AP-side NN given the input data $\bfx_{\text{AP}}\in\bbR^{d_{\bfc}}$ is defined as $\netAP(\bfx_{\text{AP}},\parAP)$. Then, the output of the overall NN is represented by $\netWH(\bfx_{\text{CL}},\bsth)=\netAP(\netUE(\bfx_{\text{CL}},\parUE),\parAP)$.

Mini-batch SGD is adopted on both the client and the AP sides for NN parameter updates.  Since the training NN is split across the client and the AP, essential information must be exchanged between them during the forward propagation and backpropagation steps. In forward propagation, since the activation at the cut layer serves as the output of the client-side NN and the input to the AP-side NN, simultaneously, these values must be transmitted from the client to the AP. Similarly, in backpropagation, since the gradient of the loss with respect to the cut-layer activation serves as the output of the AP-side NN and the input to the client-side NN, these gradient values must be transmitted from the AP to the client. 

We denote by $N\in\mathbb{Z}^+$ the maximum number of malicious clients that the system can tolerate while preserving learning robustness. The AP’s goal is to train the global and local model parameters while mitigating the disruptive impact of up to $N$ malicious participants. Such adversaries may conceal their identity by altering the information they exchange during training. Throughout this article, we consider three critical attacks:
\begin{itemize}
  \item \emph{Label Flipping}: Malicious clients intentionally alter local labels before sending them to the AP, leading to misclassification and corrupted model updates.
  
  \item \emph{Activation Tampering}: During the forward pass, distorted cut-layer activations are sent by malicious clients, degrading feature representation and overall model performance.
  
  \item \emph{Gradient Tampering}: In backpropagation, malicious clients manipulate received gradients before updating local models, steering the global model toward suboptimal convergence.
\end{itemize}

\begin{figure*}[t]
    \centering
    \subfloat{\includegraphics[width=0.7\textwidth]{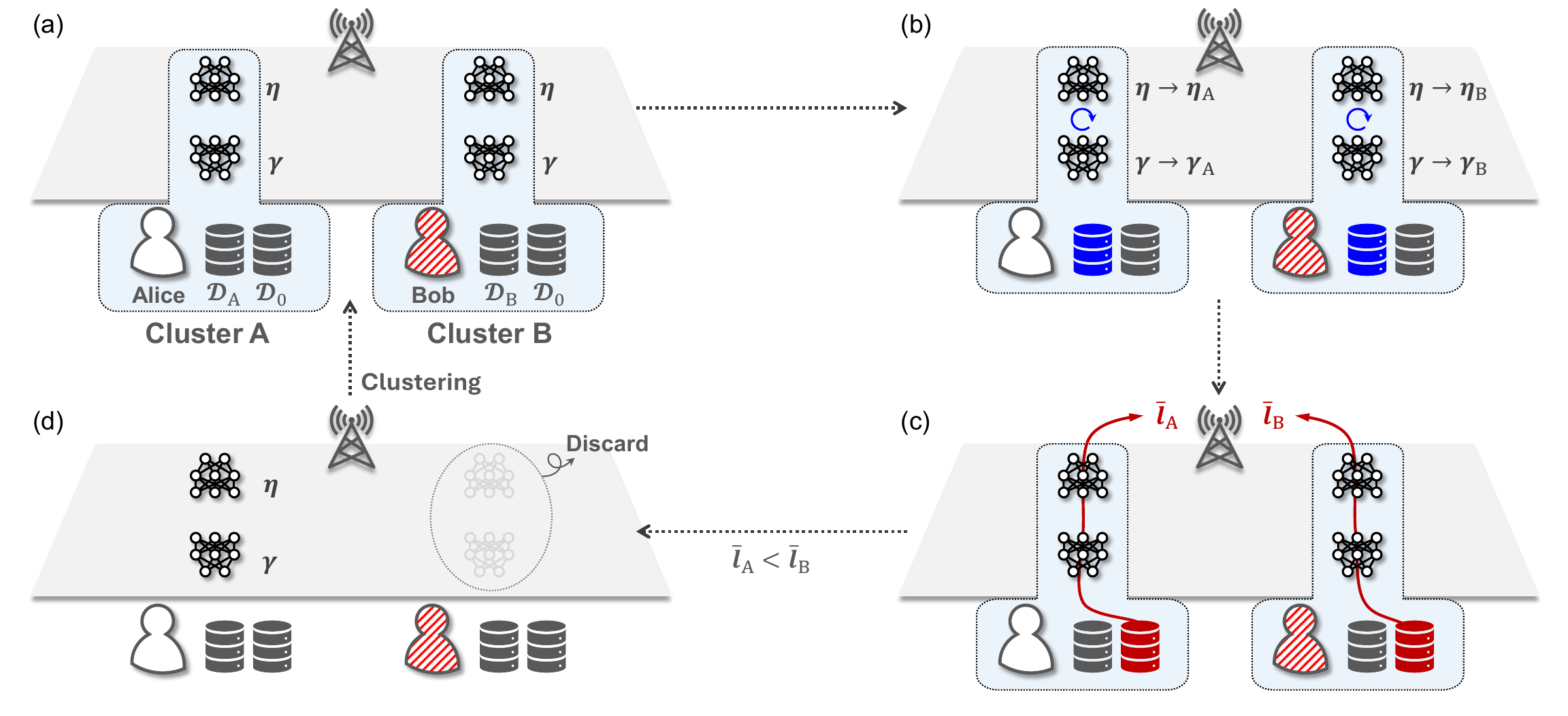}} 
    \caption{Pigeon-SL with $M=2$ and $N=1$: (a) cluster formation, (b) cluster-wise training, (c) loss evaluation, and (d) cluster selection
    }
    \label{fig:simple}
\end{figure*}

\section{Pigeonhole Principle-based SL}
\subsection{Illustrative Example}

Consider the simplest setting of our scheme, shown in Figure~\ref{fig:simple}.  Suppose there are two clients ($M=2$), Alice and Bob, and we wish to tolerate up to one malicious client ($N=1$).  Suppose Bob behaves maliciously.

\begin{enumerate}
  \item \textit{Cluster Formation} (Fig.~\ref{fig:simple}a):  
  Clients are split into two clusters—Cluster A with Alice and Cluster B with Bob. By the pigeonhole principle, at most one cluster can be malicious, ensuring at least one honest cluster. Both clusters are initialized with identical client-side ($\parUE$) and AP-side ($\parAP$) parameters.

  \item \textit{Cluster-wise Training} (Fig.~\ref{fig:simple}b):  
  Each cluster performs vanilla SL independently. Alice updates Cluster A’s parameters $(\parUE_A,\parAP_A)$ using her local data $\calD_A$, while Bob, acting maliciously, updates Cluster B’s parameters $(\parUE_B,\parAP_B)$ on $\calD_B$, possibly in an adversarial manner.

  \item \textit{Loss Evaluation} (Fig.~\ref{fig:simple}c):  
  The AP evaluates each cluster using a shared validation set $\calD_0$, computing losses $\bar{\ell}_A$ and $\bar{\ell}_B$. Typically, $\bar{\ell}_A < \bar{\ell}_B$ due to Alice’s honest updates. If $\bar{\ell}_B$ is lower, it simply indicates better performance, regardless of intent.

  \item \textit{Cluster Selection} (Fig.~\ref{fig:simple}d):  
  The AP selects the cluster with the lower validation loss (e.g., Cluster A) and discards the other. The selected parameters initialize the next training round.
\end{enumerate}

This toy example illustrates the core idea: by training $N+1$ clusters in parallel and selecting the best-performing one, the system ensures that only honest updates are retained, even when up to $N$ clients are malicious. The following sections extend this mechanism to the general client size and arbitrary adversary bound.


\subsection{Learning Process in Pigeon-SL}
At the start of the $t$‑th global round ($t\in[T]$), the AP randomly partitions the $M$ clients into $R=N+1$ clusters 
$\mathcal{Q}_r^t\subset[M]$ for $r\in[R]$, each of size $\bar{M}=M/R\in\mathbb{Z}^+$. By the pigeonhole principle, even if up to $N$ clients are malicious, at least one cluster must contain only honest participants.  
The clusters satisfy  
\begin{align}
  \text{(i)}\quad \mathcal{Q}_r^t \cap \mathcal{Q}_{r'}^t = \emptyset,
  \quad
  \text{(ii)}\quad \bigcup_{r=1}^{R} \mathcal{Q}_r^t = [M].
  \label{eq:cluster_cond}
\end{align}

Let $\bsth^{t}=[\parUE^{t},\parAP^{t}]\in\bbR^d$ denote the original network parameters before round $t$, where $\parUE^{t}\in\bbR^{d_{\text{CL}}}$ and $\parAP^{t}\in\bbR^{d_{\text{AP}}}$ are the client‑side and AP‑side parameters, respectively. Corresponding to each cluster, there is a separate split‑learning network; the AP and the $\bar{M}$ clients in cluster $\mathcal{Q}_r^t=\{\ic_{r,1}^t,\dots,\ic_{r,\bar{M}}^t\}$ collaboratively train that network using vanilla SL \cite{SL_conv}.  Within cluster $r$, updates proceed sequentially from client $\ic_{r,1}^t$ through client $\ic_{r,\bar{M}}^t$, producing updated parameters for that cluster in each training round.

During the parameter update for client $\ic_{r,{\bar{m}}}^t\in\calQ_r^t$, the learning process involves the transmission of the outputs of forward and backward propagation between the client and the AP. Specifically, at the start of mini-batch iteration $e\in[E]$, the client-side and AP-side training parameters are denoted as $\parUE_{r,\bar{m}}^{t,e}\in\bbR^{d_{\text{CL}}}$ and $\parAP_{r,\bar{m}}^{t,e}\in\bbR^{d_{\text{AP}}}$, respectively, where the parameters of client-side and AP-side NNs are initialized as $\parUE_{r,1}^{t,1}= \parUE^{t}$ and $\parAP_{r,1}^{t,1}=\parAP^{t}$, respectively. The concatenated representation of these parameters is denoted as $\bsth_{{r,{\bar{m}}}}^{t,e}=[\parUE_{r,{\bar{m}}}^{t,e}, \parAP_{r,{\bar{m}}}^{t,e}]\in\bbR^d$.

The client selects a mini-batch $\calB_{r,{\bar{m}}}^{t,e}\subset\calD_{q_{r,{\bar{m}}}^t}$ with a size of $B=\lvert\calB_{r,{\bar{m}}}^{t,e}\rvert$. Then, the gradients at both the client and the AP sides for updating the model parameters are obtained through the following steps of forward and backward propagation as in vanilla SL:
\begin{enumerate}   
    \item Client $\ic_{r,{\bar{m}}}^t$ takes the input batch $\calB_{r,{\bar{m}}}^{t,e}$ and passes it through the client-side NN. Then, it transmits the activation outputs of the cut layer, denoted as $\netUE(\bfx_b,\parUE_{r,{\bar{m}}}^{t,e})\in\bbR^{d_\bfc}$, to the AP, along with the corresponding label $y_b$ for each sample $\bfs_b=(\bfx_b,y_b)\in\calB_{r,{\bar{m}}}^{t,e}$. 
    \item After receiving the activation outputs from the client, the AP processes them through the AP-side NN to obtain the outputs of the original NN, represented as $\netAP(\netUE(\bfx_b,\parUE_{r,\bar{m}}^{t,e}),\parAP_{r,\bar{m}}^{t,e})$. Then, it computes the loss functions based on these outputs and the received label $y_b$.
    \item After performing backward gradient computation, the AP obtains the gradients for all layers in the AP-side NN. Then, it transmits the gradients of the cut layer, denoted as $\nabla_{\!\bfc}\ell(\bfs_b;\bsth_{r,\bar{m}}^{t,e})\in\bbR^{d_\bfc}$ for each sample $\bfs_b\in\calB_{r,\bar{m}}^{t,e}$, back to client $\ic_{r,{\bar{m}}}^t$.
    \item The received gradients are then propagated backward through the client-side NN, enabling the client to obtain the gradients for all its layers.
\end{enumerate}

After this process, both client $\ic_{r,{\bar{m}}}^t$ and the AP obtain the gradients for all layers in their respective NNs. Consequently, the client and the AP update their parameters $\parUE_{r,{\bar{m}}}^{t,e}$ and $\parAP_{r,{\bar{m}}}^{t,e}$ as 
\begin{align}
    \parUE_{r,{\bar{m}}}^{t,e+1}=\parUE_{r,{\bar{m}}}^{t,e} - \lambda \nabla_{\!\parUE} \bar{\ell}_{r,{\bar{m}}}^{t,e+1} \nonumber \\
    \parAP_{r,{\bar{m}}}^{t,e+1}=\parAP_{r,{\bar{m}}}^{t,e} - \lambda \nabla_{\!\parAP} \bar{\ell}_{r,{\bar{m}}}^{t,e+1} \nonumber
\end{align}
where $\lambda>0$ is a learning rate and
\begin{align}
    \nabla_{\!\bfp} \bar{\ell}_{r,{\bar{m}}}^{t,e+1}=\frac{1}{B}\sum_{\bfs_b\in\calB_{r,{\bar{m}}}^{t,e}}\nabla_{\!\bfp}\ell(\bfs_b;\bsth_{r,{\bar{m}}}^{t,e}),\ \bfp\in\{\parUE,\parAP\}
\end{align}
represents the mini-batch averaged gradients of the client-side and AP-side parameters. 

After completion of $E$ mini-batch updates, the update turn is passed from client $\ic_{r,{\bar{m}}}^t$ to client $\ic_{r,{\bar{m}+1}}^t$. The parameters of client $\ic_{r,{\bar{m}+1}}^t$ are then newly assigned as $\bsth_{r,{\bar{m}+1}}^{t,1}=\bsth_{r,{\bar{m}}}^{t,E+1}$, and the aforementioned learning process is repeated iteratively.

After the last update of the client, the parameters at the end of the $t$-th global round for cluster $\calQ_r^t$ are given by $\bsth_r^{t+1}=\bsth_{r,\bar{M}}^{t,E+1}$. Then, $\calQ_{\hat{r}}^t$ is selected based on the cluster selection criterion elaborated in the next subsection, and the parameters at the $(t+1)$-th global round are denoted as $\bsth^{t+1} = \bsth_{\hat{r}}^{t+1}$.
At the $(t+1)$-th global epoch, the AP regenerates the clusters. Then, the last client $q_{\hat{r},{\bar{M}}}^t$ in the selected cluster shares its updated parameters with all first clients $q_{r,1}^{t+1}\in\calQ_{r}^{t+1}$ for $r\in[R]$, and the aforementioned learning process is executed repeatedly. The whole training process is described in \textbf{Algorithm \ref{alg:alg1}}.

\subsection{Finding a Pigeonhole without Malicious Clients}
As observed in the previous learning process, malicious clients can disrupt updates by distorting exchanged information. In fact, the AP cannot distinguish between normal and malicious clients. Therefore, it is necessary to develop a novel scheme that defends against harmful attacks without directly identifying malicious clients. Our proposed scheme mitigates their impact by comparing and evaluating the effectiveness of the acquired parameters for each cluster during training.

\begin{algorithm}[t]
\caption{Pigeon-SL}\label{alg:alg1}
{\small
\begin{algorithmic}[1]
\REQUIRE{$\bsth^1$}
\ENSURE $\bsth^{T+1}$
\FOR{$t\gets1$ \TO $T$}
    \STATE Generate $R$ clusters as \eqref{eq:cluster_cond}
    \FOR{$r\gets1$ \TO $R$}
        \FOR{$\bar{m}\gets1$ \TO $\bar{M}$}
            \IF{$\bar{m}=1$}
                \STATE $\bsth_{r,1}^{t,1}\gets\bsth^{t}$
            \ELSE
                \STATE $\bsth_{r,\bar{m}}^{t,1}\gets\bsth_{r,\bar{m}-1}^{t,E+1}$
            \ENDIF
            \FOR{$e\gets1$ \TO $E$}
                \STATE $\calB_{r,\bar{m}}^{t,e}\gets$ (batch of size $B$ from $\calD_{q_{r,\bar{m}}^t}$)
                \FOR{$\bfs_b=(\bfx_b,y_b)\in\calB_{r,\bar{m}}^{t,e}$}
                    \STATE \textbf{FwdProp($q_{r,\bar{m}}^t,\bfs_b,\bsth_{r,\bar{m}}^{t,e}$)}            
                    \STATE \textbf{BackProp($q_{r,\bar{m}}^t,\bfs_b,\bsth_{r,\bar{m}}^{t,e}$)}
                \ENDFOR
                \STATE $\nabla_{\!\bsth}\bar{\ell}_{r,\bar{m}}^{t,e}\gets\frac{1}{B}\sum_{\bfs_b\in\calB_{r,\bar{m}}^{t,e}}\nabla \ell(\bfs_b;\bsth_{r,\bar{m}}^{t,e})$
                \STATE $\bsth_{r,\bar{m}}^{t,e+1}\gets\bsth_{r,\bar{m}}^{t,e}-\lambda\nabla \bar{\ell}_{r,\bar{m}}^{t,e}$
            \ENDFOR
        \ENDFOR
        \STATE $\bsth_r^{t+1}\gets \bsth_{r,\bar{M}}^{t,E+1}$
        \FOR{$\bfs_0=(\bfx_0,y_0)\in\calD_o$}
            \STATE \textbf{FwdProp($q_{r,\bar{M}}^t,\bfs_0,\bsth_r^{t+1}$)}
        \ENDFOR
        \STATE $\bar{\ell}_r^t\gets\frac{1}{|\calD_o|}\sum_{\bfs_0\in\calD_o}\ell(\bfs_0;\bsth_r^{t+1})$
    \ENDFOR
    \STATE $\hat{r}\gets {\arg\min}_{r\in[1:R]}\bar{\ell}_{r}^t$ 
    \STATE $\bsth^{t+1}\gets \bsth_{\hat{r}}^{t+1}$
\ENDFOR
\end{algorithmic}
}
\label{alg1}
\end{algorithm}

\begin{algorithm}[t]
    \caption{\textbf{FwdProp}($m,\bfs=(\bfx,y),\bsth=[\parUE,\parAP]$)}\label{alg:alg2}
    {\small
    \begin{algorithmic}[1]
        \ENSURE Storing all activation values of layers
        \STATE \textbf{At client $m$:}
        \STATE \hspace{\algorithmicindent}Forward propagation with $\bfx$ to get $\netUE(\bfx,\parUE)$
        \STATE \hspace{\algorithmicindent}Transmitting $\netUE(\bfx,\parUE)$ and $y$ to the AP
        \STATE \textbf{At the AP:}
        \STATE \hspace{\algorithmicindent}Forward propagation with $\netUE(\bfx,\parUE)$ to get $\netAP(\netUE(\bfx,\parUE),\parAP)$
        \STATE \hspace{\algorithmicindent}Computing loss $\ell(\bfs;\bsth)$ with $y$ and $\netAP(\netUE(\bfx,\parUE),\parAP)$
    \end{algorithmic}
}
\end{algorithm}

\begin{algorithm}[t]
    \caption{\textbf{BackProp}($m,\bfs,\bsth=[\parUE,\parAP]$)}\label{alg:alg3}
    {\small
    \begin{algorithmic}[1]
        \ENSURE Storing all gradients of layers
        \STATE \textbf{At the AP:}
        \STATE \hspace{\algorithmicindent}Backpropagation to get $\nabla_{\!\parAP}\ell(\bfs;\bsth)$ and $\nabla_{\!\bfc}\ell(\bfs;\bsth)$
        \STATE \hspace{\algorithmicindent}Transmitting $\nabla_{\!\bfc}\ell(\bfs;\bsth)$ to client $m$
        \STATE \textbf{At client $m$:}
        \STATE \hspace{\algorithmicindent}Backpropagation with $\nabla_{\!\bfc}\ell(\bfs;\bsth)$ to get $\nabla_{\!\parUE}\ell(\bfs;\bsth)$
    \end{algorithmic}
}
\end{algorithm}


Basically, at the end of each global round, the last client $q_{r,\bar{M}}^t\in\calQ_r^t$ computes and transmits $\netUE(\bfx_0,\parUE_{r,\bar{M}}^{t,E+1})$ to the AP along with the labels $y_0$ for the shared samples $(\bfx_0,y_0)\in\calD_o$. Then, the AP applies each received outputs to the AP-side NN to obtain $\netWH(\bfx_0,\bsth_r^{t+1})$. Subsequently, the AP obtains the validation loss value of cluster $\calQ_r^t$ after the $t$-th round denoted as
\begin{align}
    \bar{\ell}_{r}^t=\frac{1}{|\calD_o|}\sum_{\bfs_0\in\calD_o}\ell(\bfs_0;\bsth_{r}^{t+1}). \nonumber
\end{align}
This loss value is computed by averaging the losses over the shared samples to assess the effectiveness of the trained parameters $\bsth_{r}^{t+1}$. To identify the cluster that appears to have learned most effectively, the AP compares the validation loss functions of all clusters and selects the cluster with the lowest value, denoted as 
\begin{align}
    \calQ_{\hat{r}}^t,~\hat{r}=\underset{r\in[R]}{\arg\min}~\bar{\ell}_{r}^t. \nonumber
\end{align}

Validation loss values of clusters implicitly reflect the average learning performance of their clients. Malicious clients typically exhibit higher losses, raising the overall cluster loss. Thus, selecting the cluster with the lowest validation loss serves as an effective strategy to enhance robustness by leveraging benign client contributions.

By setting the total number of clusters $R$ to be one more than $N$, the maximum number of malicious clients, the system ensures that at least one cluster consists solely of benign clients, even in the worst case. Consequently, it is highly likely that at least one cluster yields a low validation loss. Even if a mixed cluster shows lower loss than a purely benign one, this suggests that benign clients in the cluster dominate learning and mitigate adversarial impact, preserving model performance.

Certainly, one might naturally question the potential threat posed by a malicious client positioned at the last update turn, which could disrupt the entire learning process by reducing the validation loss while manipulating the trained parameters. Specifically, consider the scenario that the last client $\ic_{r,\bar{M}}^t$ in cluster $\calQ_r^t$ is malicious. After its local update, the malicious client transmits the cut-layer activation values $\netUE(\bfx_0,{\parUE}_{r,\bar{M}}^{t,E+1})$ for $(\bfx_0,y)\in\calD_o$ without manipulation as usual, aiming to reduce the validation loss so that its cluster can be selected for the global update. When its cluster is selected, the malicious client transmits manipulated parameters, $\tilde{\parUE}_{r,\bar{M}}^{t,E+1}$, to the first clients in the next global round clusters instead of the well-trained ones, thereby degrading training quality. 

Fortunately, this threat can be detected in advance by the first clients in the next global round clusters. Before conducting their local updates, they transmit the cut-layer activation values $\netUE(\bfx_0,\tilde{\parUE}_{r,\bar{M}}^{t,E+1})$ for $(\bfx_0,y)\in\calD_o$. Since there are $N+1$ clients, at least one is benign and transmits its activation values without manipulation. Consequently, the AP can compare the values transmitted by the malicious and benign clients and observe a difference: $\netUE(\bfx_0,{\parUE}_{r,\bar{M}}^{t,E+1})\neq \netUE(\bfx_0,\tilde{\parUE}_{r,\bar{M}}^{t,E+1})$ for $(\bfx_0,y)\in\calD_o$. 
When the AP detects this manipulation, it discards the tampered cluster and returns to the previous global round to reselect an alternative cluster instead. As a result, malicious clients deliberately avoid manipulating the parameters to ensure that their influence is not completely nullified at once.

\subsection{Pigeon-SL+: Enhancing Efficiency and Security}

Thus far, we have described the Pigeon-SL process and its core mechanism.  While Pigeon-SL offers more stable convergence than standard vanilla SL, it updates only one cluster per round, reducing the total number of client updates by a factor of $R$ relative to vanilla SL.  Consequently, Pigeon-SL’s training pace slows—particularly when $N$ is large but few (or no) clients are actually malicious, leading to reduced efficiency compared to vanilla SL.

To overcome this limitation, we propose an enhanced algorithm, Pigeon-SL+.  Up through cluster selection based on validation loss, Pigeon-SL+ follows Pigeon-SL exactly.  Once the best cluster is chosen, however, the algorithm repeats training on that cluster for an additional $R-1$ iterations.  Including the initial iteration used for selection, the selected cluster thus undergoes $R$ rounds of training.  Since each round involves $\bar{M}$ sequential client updates, the total number of updates becomes $R \times \bar{M} = M$,
matching the per‑round update count of standard vanilla SL.  In this way, Pigeon-SL+ restores training efficiency while preserving the robustness of the original Pigeon-SL scheme.

\begin{table*}[h]
\centering
\caption{Communication and Computation Overheads of vanilla SL and proposed methods per Global Epoch \label{tab:overhead}}
\footnotesize
\begin{tabular}{lcc}
\hline
\textbf{Configuration} & \textbf{Communication Overhead} & \textbf{Computation Overhead} \\
Selected Cluster (Pigeon-SL) &
$(\bar{M} \tilde{D} +2D_o) d_\bfc + (\bar{M} + R - 1)d_{\text{CL}}$ &
$(\bar{M} \tilde{D} + 2D_o) F_{\text{CL}}$ \\
Selected Cluster (Pigeon-SL+) &
$(M \tilde{D} +2D_o) d_\bfc + (M+R-1)d_{\text{CL}}$ &
$(M \tilde{D} + 2D_o) F_{\text{CL}}$ \\
Unselected Cluster (Pigeon-SL) &
$(\bar{M} \tilde{D} +2D_o) d_\bfc + (\bar{M}- 1)d_{\text{CL}}$ &
$(\bar{M} \tilde{D} + 2D_o) F_{\text{CL}}$ \\
Unselected Cluster (Pigeon-SL+) &
$(\bar{M} \tilde{D} +2D_o) d_\bfc + (\bar{M}- 1)d_{\text{CL}}$ &
$(\bar{M} \tilde{D} + 2D_o) F_{\text{CL}}$ \\
\hline 
Total Clients (vanilla SL) & $M\tilde{D} d_{\bfc} + {M}d_{\text{CL}}$  &  $M \tilde{D} F_{\text{CL}}$ \\
Total Clients (Pigeon-SL) & $(M\tilde{D} + 2RD_o)d_{\bfc} + M d_{\text{CL}}$ & $(M \tilde{D} + 2RD_o) F_{\text{CL}}$ 
\\ 
Total Clients (Pigeon-SL+) & $((2M - \bar{M})\tilde{D} + 2RD_o)d_{\bfc} + (2M - \bar{M})d_{\text{CL}}$ & $((2M-\bar{M})\tilde{D} + 2RD_o)F_{\text{CL}}$ \\
\hline 
\end{tabular}
\end{table*}

\begin{remark}[Complexity Analysis]
We analyze the communication and computation overheads of Pigeon-SL and its extension, Pigeon-SL+. Assuming each client has $\tilde{D}$ training samples and denoting the client-side cost of one forward and backward pass as $F_{\text{CL}}$, Table~\ref{tab:overhead} compares the overheads of vanilla SL, Pigeon-SL, and Pigeon-SL+. While our schemes incur moderate overhead increases relative to vanilla SL, they offer a favorable trade-off for improved robustness. The overheads scale with $d_{\bfc}$, $d_{\text{CL}}$, and $F_{\text{CL}}$, making the proposed methods more efficient than FL-based approaches when the client-side model is compact.
\end{remark}

\section{Convergence Analysis of Pigeon-SL}
In this section, to validate the convergence of Pigeon-SL and Pigeon-SL+, we conduct a mathematical convergence analysis. As a matter of fact, since the convergence analysis of Pigeon-SL+ can be carried out in a similar manner to that of Pigeon-SL, we omit it in this paper. Firstly, we present some assumptions which are widely adopted in the literature.

\noindent \textbf{(A1)} 
$L_m$ is a $\kappa$-smooth function, i.e. for all $\bsa,\bsb\in\bbR^d$,
        \begin{align}
            \lVert\nabla L_m(\bsa)-\nabla L_m(\bsb) \rVert\le \kappa \lVert \bsa-\bsb \rVert, \label{eq:smooth}
        \end{align}
        or, equivalently,
        \begin{align}
            \lvert L_m(\bsb)-L_m(\bsa)-\nabla L_m(\bsa)^\mathsf{T}(\bsb-\bsa)\rvert
            \le \frac{\kappa}{2}\rVert\bsb-\bsa\lVert^2. \label{eq:smooth2}
        \end{align}

\noindent \textbf{(A2)} 
 For $m\in\calM$ and $\bfs\sim p_{m}$, 
        \begin{align}
            \bbE\left[{\nabla l(\bfs;\bsth)}|\bsth\right]=\nabla L_m(\bsth). \label{eq:sgd_property}
        \end{align}

\noindent \textbf{(A3)} 
For $m\in\calM$ and $\bfs\sim p_{m}$, there is $\sigma^2$ such that
        \begin{align}
            \bbE\left[{\lVert\nabla l(\bfs;\bsth) - \nabla L_m(\bsth)\rVert^2}\right]\le \sigma^2. \label{eq:boundness2}
        \end{align}

\noindent \textbf{(A4)} 
There exist $\delta^2$ such that
        \begin{align}
            \frac{1}{M}\sum_{m=1}^{M}\lVert \nabla L_m(\bsth) - \nabla L(\bsth) \rVert^2 \le \delta^2. \label{eq:diversity}
        \end{align}

Based on these assumptions, we can derive the following lemma.
\begin{lemma} \label{lem:main}
    For \textbf{Algorithm 1}, when $\lambda<\frac{1}{12\kappa\bar{M}E}$, 
    \begin{align}
        &\bbE[L(\bsth^{t+1})] - L(\bsth^t) \le -\frac{\lambda \bar{M}E}{12} \lVert  L(\bsth^t)\rVert^2 \nonumber \\
    &\qquad\qquad+6\lambda^3\kappa^2 M\bar{M}^3 E^3 \delta^2 + 6\lambda^3 \kappa^2 \bar{M}^2 E^2 \sigma^2 \nonumber \\ 
    &\qquad\qquad+ \lambda^2\kappa \bar{M}^2E^2\sigma^2 + 3\lambda^2 \kappa M \bar{M}^2E^2\delta^2.\nonumber      
    \end{align}
\end{lemma}
\begin{IEEEproof}
    See Appendix B.
\end{IEEEproof}

Finally, based on the above lemma, we can reach the following theorem showing convergence of the proposed scheme.
\begin{theorem} \label{thm:main}
    For $\bsth^t$ in \textbf{Algorithm 1} and $T\in\bbZ^+$, when $\lambda\le\frac{1}{12\kappa \bar{M}E}$,
    \begin{align}
        &\frac{1}{T} \sum_{t=1}^{T} \bbE[\lVert L(\bsth^t) \rVert^2] 
    \le -\frac{12}{\lambda T\bar{M}E} \big(\bbE[L(\bsth^{T+1})] - L(\bsth^1)\big) \nonumber \\
    &\qquad+72\lambda^2 \kappa^2 M \bar{M}^2 E^2 \delta^2 + 72 \lambda^2 \kappa^2 \bar{M} E \sigma^2 \nonumber \\
    &\qquad+12\lambda \kappa \bar{M} E \sigma^2 + 36\lambda \kappa M \bar{M} E \delta^2.
    \end{align}
    Furthermore, when $\lambda=\frac{1}{12\kappa\bar{M}E\sqrt{T}}$,
    \begin{align}
        &\frac{1}{T} \sum_{t=1}^{T} \bbE[\lVert L(\bsth^t) \rVert^2] 
    \le -\frac{144\kappa}{\sqrt{T}} \big(\bbE[L(\bsth^{T+1})] - L(\bsth^1)\big) \nonumber \\
    &\quad+\frac{M  \delta^2}{2T} +\frac{R\sigma^2}{2MET} +\frac{\sigma^2}{\sqrt{T}} + \frac{3M\delta^2}{\sqrt{T}}=\calO\bigg(\frac{1}{\sqrt{T}}\bigg).
    \end{align}
\end{theorem}
\begin{IEEEproof}
    See Appendix C.
\end{IEEEproof}
\begin{remark}[Convergence Bound Interpretation]
The expected squared gradient norm in Theorem~\ref{thm:main} is a standard metric for evaluating convergence in non-convex optimization \cite{ncvx}. Notably, the bound includes a clustering-related term $R$, which does not appear in conventional analyses. As the number of clusters increases, $R$ grows, potentially slowing convergence—reflecting the intuition that more clusters reduce the number of clients trained per epoch, delaying progress. Despite this, the convergence rate remains $\mathcal{O}(1/\sqrt{T})$, consistent with recent results in distributed and federated learning \cite{cvg_rate01,cvg_rate02,cvg_rate03}. This confirms the suitability of Pigeon-SL for modern distributed learning settings.
\end{remark}

\section{Numerical Results}

\begin{figure*}[t]
    \centering
    \subfloat[Label flipping]{\includegraphics[width=0.33\textwidth]{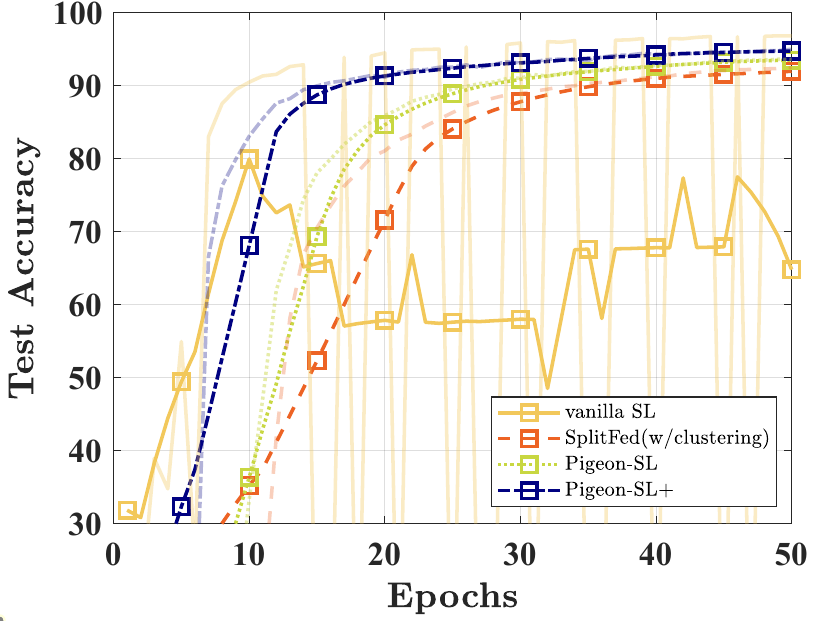}\label{fig:MNIST_LF_N_3}} 
    \subfloat[Activation tampering]{\includegraphics[width=0.33\textwidth]{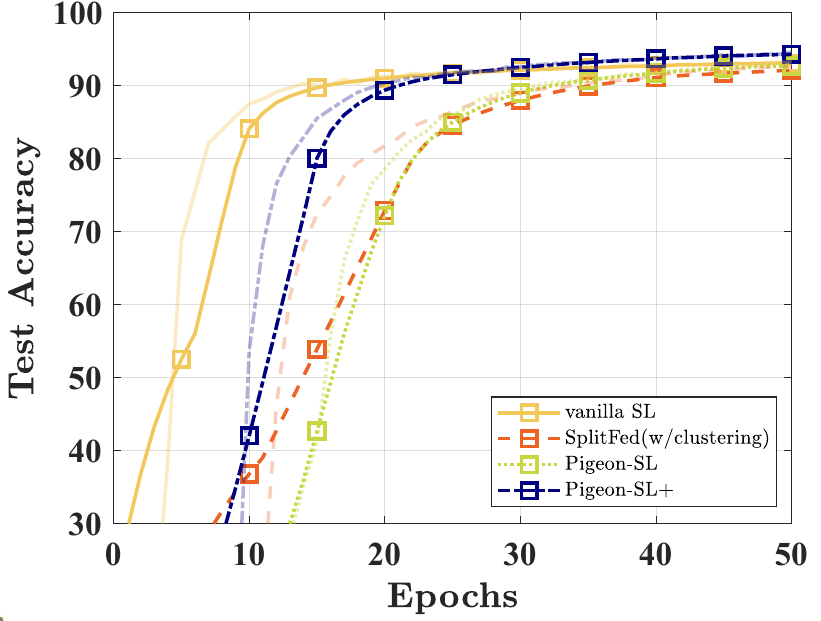}\label{fig:MNIST_AT_N_3}} 
    \subfloat[Gradient tampering]{\includegraphics[width=0.33\textwidth]{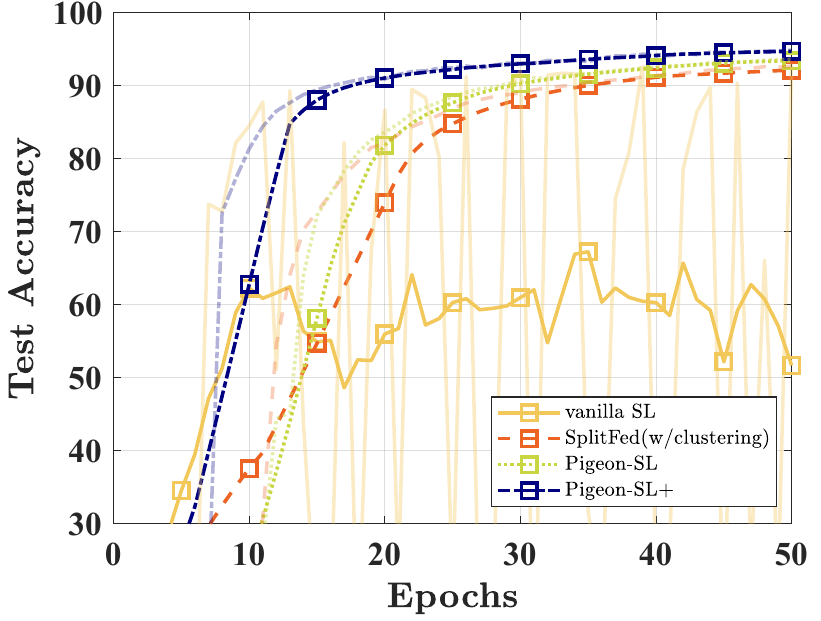}\label{fig:MNIST_GT_N_3}}
    \caption{Test accuracy comparison of MNIST classifiers between the conventional vanilla SL, SplitFed, the proposed Pigeon-SL and the enhanced Pigeon-SL+ for $N=3$.
    \vspace{-10.0pt}}
    \label{fig:MNIST_TA_N_3}
\end{figure*}

\begin{figure*}[t]
    \centering
    \subfloat[Label flipping]{\includegraphics[width=0.33\textwidth]{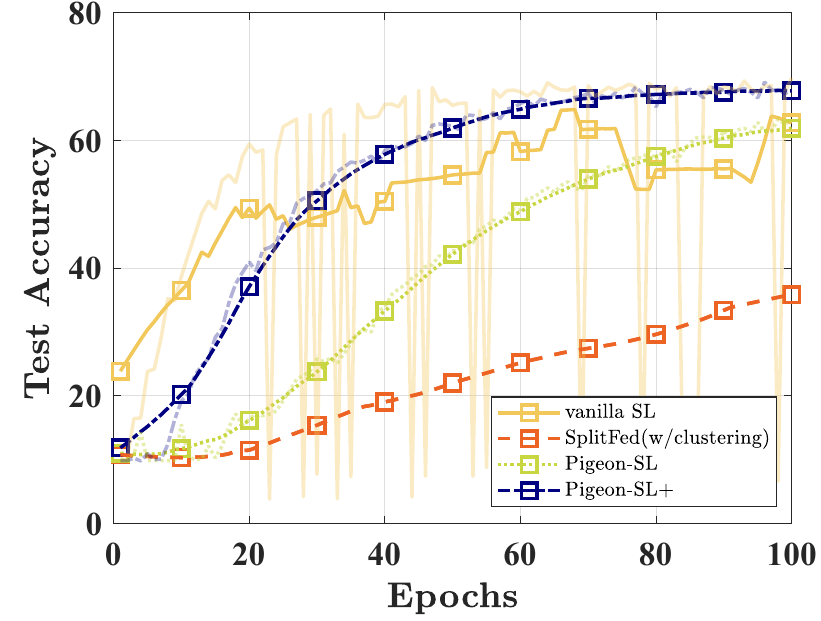}\label{fig:TA_LD_N_4}} 
    \subfloat[Activation tampering]{\includegraphics[width=0.33\textwidth]{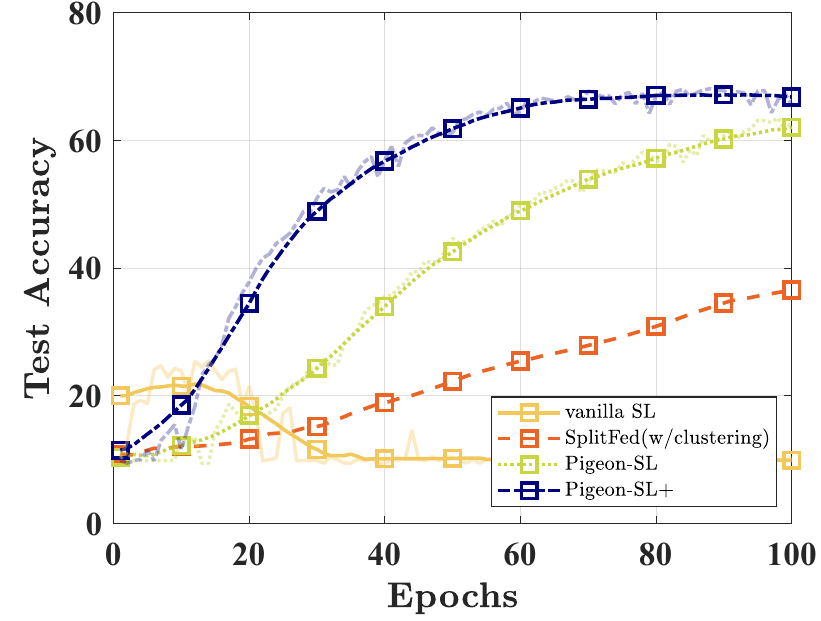}\label{fig:TA_AD_N_4}} 
    \subfloat[Gradient tampering]{\includegraphics[width=0.33\textwidth]{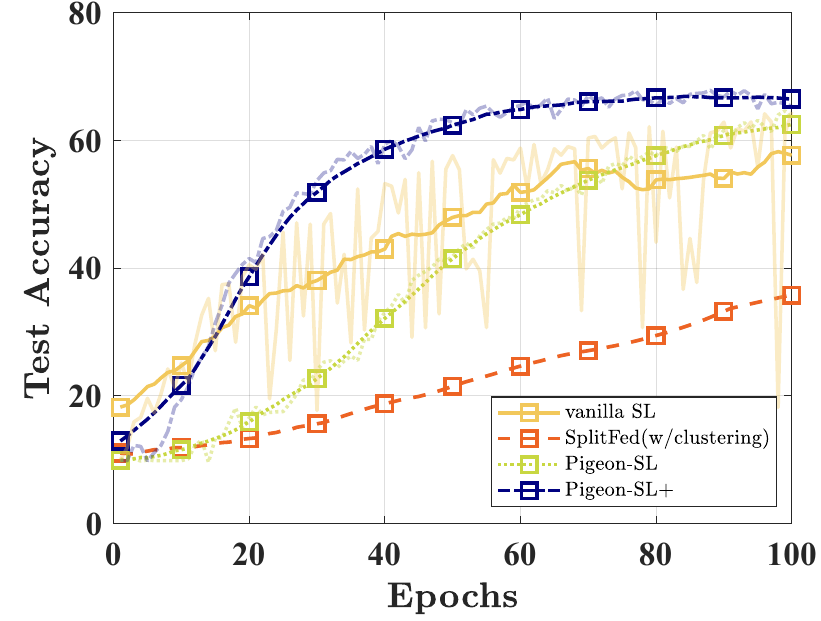}\label{fig:TA_GD_N_4}}
    \caption{Test accuracy of CIFAR-10 classifiers comparison between the conventional vanilla SL, SplitFed, the proposed Pigeon-SL and the enhanced Pigeon-SL+ for $N=4$.
    \vspace{-10.0pt}}
    \label{fig:CIFAR_TA_N_4}
\end{figure*}

\begin{figure*}[t]
    \subfloat[Label flipping]{\includegraphics[width=0.33\textwidth]{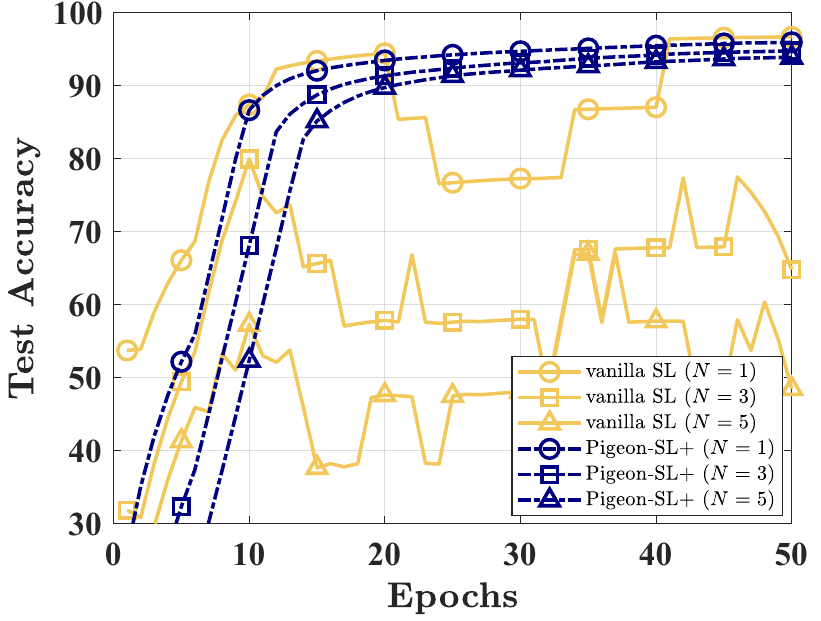}\label{fig:MNIST_LF}} 
    \subfloat[Activation tampering]{\includegraphics[width=0.33\textwidth]{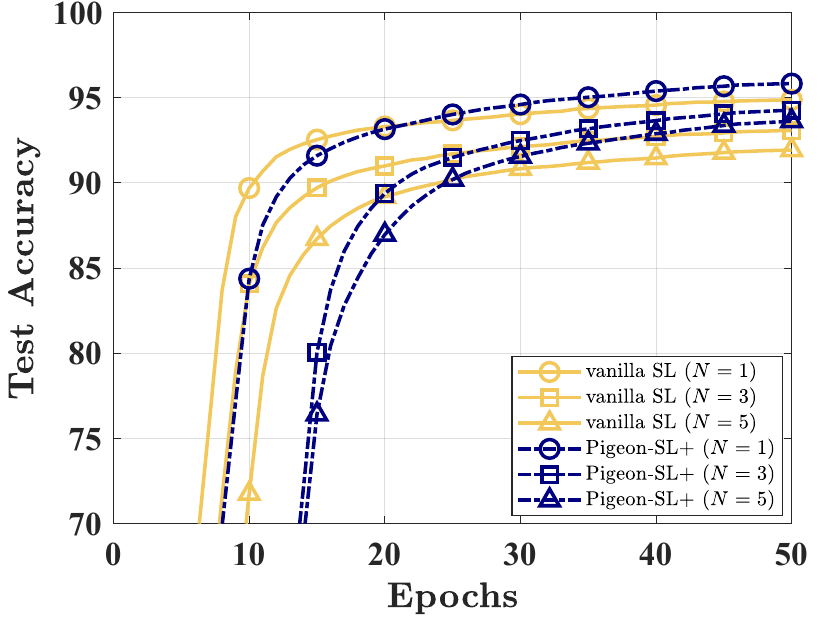}\label{fig:MNIST_AT}} 
    \subfloat[Gradient tampering]{\includegraphics[width=0.33\textwidth]{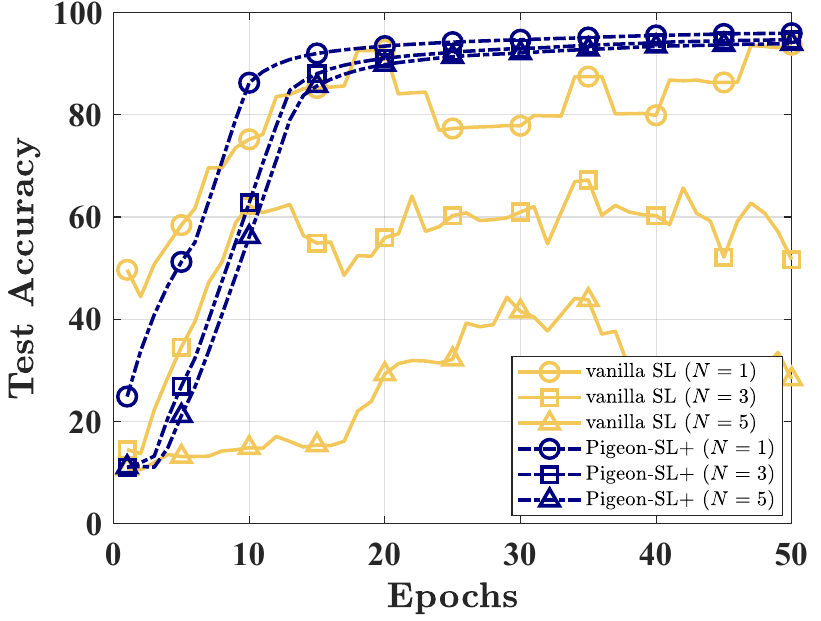}\label{fig:MNIST_GT}}
    \caption{Test accuracy comparison of MNIST classifiers between the conventional vanilla SL and the enhanced Pigeon-SL+ for various $N$.
    \vspace{-10.0pt}}
    \label{fig:MNIST_TA}
\end{figure*}

\begin{figure*}[t]
    \subfloat[Label flipping]{\includegraphics[width=0.33\textwidth]{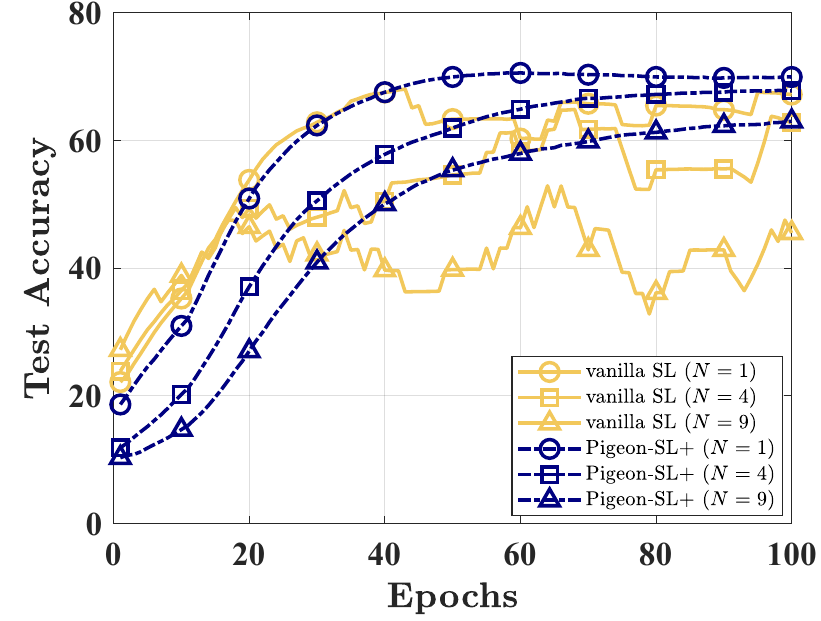}\label{fig:lab_dist_Byz}} 
    \subfloat[Activation tampering]{\includegraphics[width=0.33\textwidth]{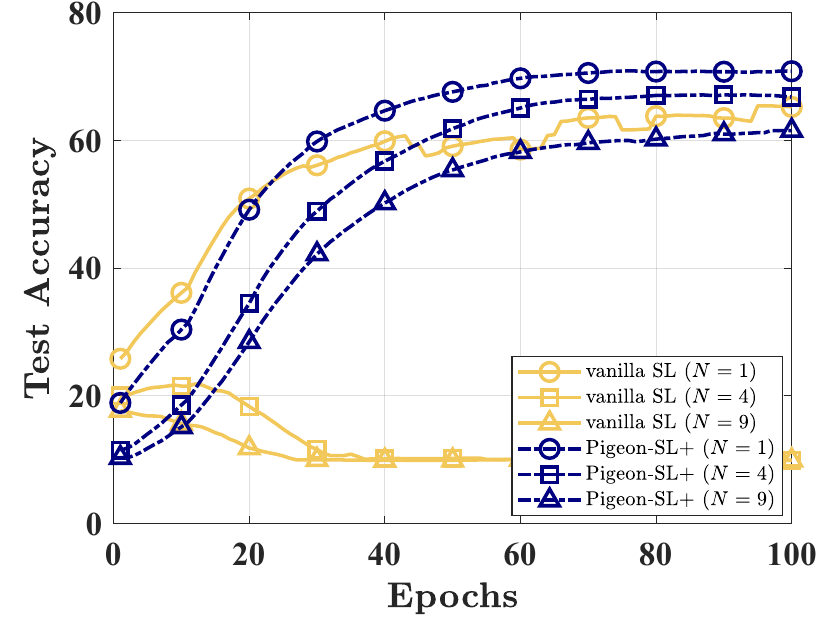}\label{fig:act_dist_Byz}} 
    \subfloat[Gradient tampering]{\includegraphics[width=0.33\textwidth]{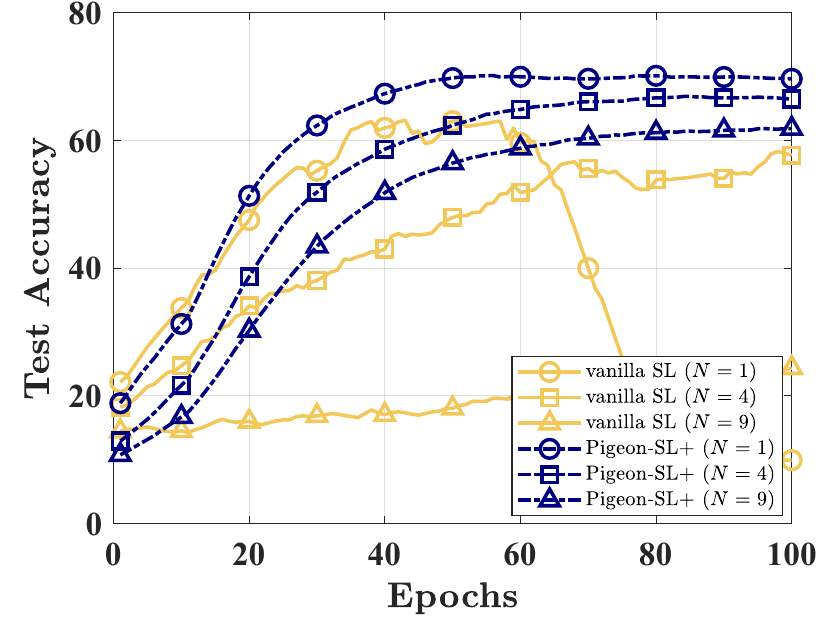}\label{fig:grad_dist_Byz}}
    \caption{Test accuracy comparison of CIFAR-10 classifiers between the conventional vanilla SL and the enhanced Pigeon-SL+ for various $N$.
    \vspace{-10.0pt}}
    \label{fig:CIFAR_TA}
\end{figure*}

\begin{table}[t]
\small
\centering
\caption{Simulation Environments}
\label{tab:sim_env}
\footnotesize
\begin{tabular}{l|cc}
\hline
\textbf{Parameters} & MNIST & CIFAR-10 \\ \hline
$D_m$ & $5,000$ & $2,500$  \\
$D_o$ & $3,000$ & $3,000$\\ 
$M$ & $12$ & $20$ \\
$B$ & $64$ & $64$ \\
$E$ & $79$   & $40$ \\
$\lambda$ & $0.001$ & $0.0002$ \\
$\lambda$ (SFL) & $0.01$ & $0.002$ \\
$R$ & $2, 4, 6$ & $2, 5, 10$ \\
$N$ (Malicious Clients) & $1,3,5$ & $1, 4, 9$ \\
\hline
\end{tabular}
\end{table}

\subsection{Simulation environment}
We implemented image classifiers on the MNIST \cite{deng2012mnist} and CIFAR-10 \cite{CIFAR10} datasets. The MNIST model uses two convolutional layers (2 and 4 filters, $5{\times}5$ kernel, padding 2), followed by a 32-node fully connected cut layer and a 10-node output layer. The CIFAR-10 model includes three convolutional layers (32, 64, 128 filters, $3{\times}3$ kernel) and four fully connected layers (256, 128, 64, 10 nodes), with the first acting as the cut layer.

We adopt the cross-entropy loss \cite{CEL} and list other parameters in Table~\ref{tab:sim_env}. For comparison, we adapt SFL \cite{splitfed} by integrating our clustering and validation-based selection, using a learning rate 10$\times$ higher to match the convergence speed of SL. FL-based methods are omitted due to incompatible attack models.

As mentioned above, the malicious client $q_{r,\bar{m}}^t$ in the network employs three attack schemes, which are described below:
\begin{itemize}
    \item Label Flipping: malicious client $q_{r,\bar{m}}^t$ corrupts the transmit labels as $y_b+3~\text{(mod $10$)}$ for each $(\bfx_b,y_b)\in\calB_{r,\bar{m}}^{t,e}$. This means the labels are shifted three positions behind the true ones.
    \item Activation Tampering: for all $(\bfx_b,y_b)\in\calB_{r,\bar{m}}^{t,e}$ in the simulations, malicious client $q_{r,\bar{m}}^t$ modifies the activation outputs from $\netUE(\bfx_b,\parUE_{r,\bar{m}}^{t,e})$ to $0.1\times\netUE(\bfx_b,\parUE_{r,\bar{m}}^{t,e}) + 0.9 \times \tilde{\mathbf{n}}$ where $\tilde{\mathbf{n}} = (\lVert\netUE(\bfx_b,\parUE_{r,\bar{m}}^{t,e})\rVert/\lVert\mathbf{n}\rVert)\cdot\mathbf{n}$ is a random vector normalized by the norm of activation outputs for a $d_{\textbf{c}}$-dimensional vector $\mathbf{n}$ generated from the standard normal distribution. 
    \item Gradient Tampering: in the simulations, malicious client $q_{r,\bar{m}}^t$ reverses the gradients at the cut layer received from the AP as $-\nabla_{\!\bfc}l(\bfx_b,y_b;\bsth_{r,\bar{m}}^{t,e})\in\bbR^{d_{\bfc}}$ for all $(\bfx_b,y_b)\in\calB_{r,\bar{m}}^{t,e}$, it performs backpropagation to update its parameters.
\end{itemize}

The performance evaluation of the classifier is based on the test accuracy, which represents the ratio of correct predictions for the 7,000 test samples. In particular, for Pigeon-SL, Pigeon-SL+, and SFL, the test accuracy is measured based on the final update from the selected cluster.

\subsection{Simulation results}
Fig. \ref{fig:MNIST_TA_N_3} and \ref{fig:CIFAR_TA_N_4}  present the test accuracy of the MNIST and CIFAR-10 classifiers of each type of attack when $N=3$ and $4$, respectively. In each figure, the yellow solid line, orange dashed line, green dotted line, and blue dash-dot line represent vanilla SL, SFL, Pigeon-SL, and Pigeon-SL+, respectively. Transparent lines without markers indicate the raw values at each round, while the bold lines with square markers indicate moving averages computed with a window size of 10 and 20, respectively. 

In the case of the MNIST classifier, When using vanilla SL, the raw test accuracy shows significant fluctuations for both label flipping and gradient tampering, indicating unstable training behavior. In contrast, stable training is observed in the other three schemes, where Pigeon-SL and Pigeon-SL+ achieve both faster convergence and better accuracy than SFL. While vanilla SL converges quickly under activation tampering, Pigeon-SL+ ultimately yields better accuracy due to its robustness against the attack.

In CIFAR-10 simulations, the contrast in performance becomes clearer. In particular, under activation tampering, Pigeon-SL and Pigeon-SL+ train successfully, whereas vanilla SL completely fails to learn. In addition, SFL trains very slowly in all cases, which shows that the proposed schemes are more efficient than other approaches.

Taking into account that Pigeon-SL+ has already shown better stability and faster convergence than Pigeon-SL and SFL, Fig. \ref{fig:MNIST_TA} and \ref{fig:CIFAR_TA} show the test accuracy of the MNIST and CIFAR-10 classifiers for various $N$ when using Pigeon-SL+ and vanilla SL. In each figure, the yellow solid line and blue dash-dot line represent vanilla SL  and Pigeon-SL+, respectively. Also, the lines with circle, square, and triangle markers represent the cases where $N=1,3,5$ and $N=1,4,9$, respectively. Each line represents values smoothed by a moving average with a window size of 10 and 20, respectively, in order to enable a fair comparison of average performance between Pigeon-SL+ and vanilla SL. Despite minor differences across figures, Pigeon-SL+ generally trains faster and more accurately than vanilla SL. Additionally, it can be observed that both test accuracy and convergence speed degrade as $N$ increases, which is consistent with our expectations.

\section{Conclusion}
This paper introduced a novel clustering-based SL scheme designed to enhance robustness against attacks from malicious clients. The proposed approach utilizes AP to compute the averaged validation loss for each cluster. By identifying the cluster with the minimal validation loss, the model is updated, minimizing the influence of malicious clients. Additionally, we proposed a tamper-proof technique based on shared samples, which prevents manipulation of the validation loss by malicious clients. Simulation results demonstrated that the proposed scheme achieved high training performance even in the presence of information distortion attacks.

\setcounter{equation}{28}
    \begin{figure*}
 {\footnotesize
        \begin{align}
            \bbE[\lVert\bsth_{r,\bar{m}}^{t,e} - \bsth^t \rVert^2] &= \lambda^2\bbE\bigg[\bigg\lVert 
    \sum_{\mu=1}^{\bar{m}}\sum_{\epsilon=1}^{\calE(\mu)}\nabla \bar{\ell}_{r,\mu}^{t,\epsilon}\bigg\rVert^2\bigg] \label{lem:pf01} \\
    &=\lambda^2\bbE\bigg[\bigg\lVert 
    \sum_{\mu=1}^{\bar{m}}\sum_{\epsilon=1}^{\calE(\mu)}\big(\nabla \bar{\ell}_{r,\mu}^{t,\epsilon} - \nabla L_{\mu}(\bsth_{r,\mu}^{t,\epsilon}) + \nabla L_{\mu}(\bsth_{r,\mu}^{t,\epsilon}) 
    - \nabla L_{\mu}(\bsth^t) + \nabla L_{\mu}(\bsth^t) - \nabla L(\bsth^t) + \nabla L(\bsth^t)\big)  \bigg\rVert^2\bigg]  \\ 
    &\le 4\lambda^2\underbrace{\bbE\bigg[\bigg\lVert 
    \sum_{\mu=1}^{\bar{m}}\sum_{\epsilon=1}^{\calE(\mu)}\big(\nabla \bar{\ell}_{r,\mu}^{t,\epsilon} - \nabla L_{\mu}(\bsth_{r,\mu}^{t,\epsilon})\big) \bigg\rVert^2 \bigg]}_{\triangleq \text{(L.1)}}
    + 4\lambda^2 \underbrace{\bbE\bigg[\bigg\lVert \sum_{\mu=1}^{\bar{m}}\sum_{\epsilon=1}^{\calE(\mu)}\big(\nabla L_{\mu}(\bsth_{r,\mu}^{t,\epsilon}) - \nabla L_{\mu}(\bsth^t) \big)\bigg\rVert^2 \bigg]}_{\triangleq\text{(L.2)}} \nonumber \\
    &\qquad\qquad\qquad\qquad\qquad+  4\lambda^2 \underbrace{\bbE\bigg[\bigg\lVert \sum_{\mu=1}^{\bar{m}}\sum_{\epsilon=1}^{\calE(\mu)} \big(\nabla L_{\mu}(\bsth^t) - \nabla L(\bsth^t)\big)\bigg\rVert^2 \bigg]}_{\triangleq\text{(L.3)}}
    +  4\lambda^2 \underbrace{\bbE\bigg[\bigg\lVert\sum_{\mu=1}^{\bar{m}}\sum_{\epsilon=1}^{\calE(\mu)}\nabla L(\bsth^t)\big)\bigg\rVert^2\bigg]}_{\triangleq \text{(L.4)}} \label{lem:pf02}
        \end{align}
        }
    \hrule    
    \end{figure*}

\setcounter{equation}{9}

\appendices
\section{Useful Lemmas with Proofs}
\begin{lemma} \label{lem:loss_exp}
    In \textbf{Algorithm 1}, the expected empirical loss $\bar{\ell}_r^t$ satisfies
    \begin{align}
        \bbE[\bar{\ell}_r^t] = \bbE[L(\bsth_r^{t+1})],
    \end{align}
    where $L(\bsth) \triangleq \frac{1}{D_o} \sum_{\bfs \in \calD_o} \ell(\bfs; \bsth)$.
\end{lemma}
\begin{IEEEproof}
    By definition of $\bar{\ell}_r^t$,
    \begin{align}
        \bbE[\bar{\ell}_r^t] 
        &= \bbE\left[\frac{1}{D_o} \sum_{\bfs \in \calD_o} \ell(\bfs; \bsth_r^{t+1}) \right]\\
        &= \frac{1}{D_o} \sum_{\bfs \in \calD_o} \bbE[\ell(\bfs; \bsth_r^{t+1})]\\
        &= \bbE[L(\bsth_r^{t+1})].
    \end{align}
\end{IEEEproof}

\begin{lemma} \label{lem:bound_Loss1}
    For any $m \in [M]$ and $\bsth \in \bbR^d$,
    \begin{align}
        \lVert \nabla L_m(\bsth) - \nabla L(\bsth) \rVert^2 \le M\delta^2.
    \end{align}
\end{lemma}
\begin{IEEEproof}
    From \textbf{(A4)}, 
    \begin{align}
        \sum_{m=1}^M \lVert \nabla L_m(\bsth) - \nabla L(\bsth) \rVert^2 \le M\delta^2,
    \end{align}
    so the claim holds for any $m \in [M]$.
\end{IEEEproof}

\begin{lemma} \label{lem:bound_Loss2}
    For $\bsth_{r,\bar{m}}^{t,e}$ in \textbf{Algorithm 1},
    \begin{align}
        \bbE\left[\left\lVert \nabla \bar{\ell}_{r,\bar{m}}^{t,e} - \nabla L_{\bar{m}}(\bsth_{r,\bar{m}}^{t,e}) \right\rVert^2\right] \le \sigma^2.
    \end{align}
\end{lemma}
\begin{IEEEproof}
    By definition of $\bar{\ell}_{r,\bar{m}}^{t,e}$,
    \begin{align}
        &\bbE\left[\left\lVert \frac{1}{B} \sum_{\bfs \in \calB_{r,\bar{m}}^{t,e}} \nabla \ell(\bfs; \bsth_{r,\bar{m}}^{t,e}) - \nabla L_{\bar{m}}(\bsth_{r,\bar{m}}^{t,e}) \right\rVert^2 \right] \\
        &= \bbE\!\left[\left\lVert \frac{1}{B} \!\sum_{\bfs \in \calB_{r,\bar{m}}^{t,e}}\!\! \left( \nabla \ell(\bfs; \bsth_{r,\bar{m}}^{t,e}) \!-\! \nabla L_{\bar{m}}(\bsth_{r,\bar{m}}^{t,e}) \right) \right\rVert^2 \right]\\
        &\le \frac{1}{B} \sum_{\bfs \in \calB_{r,\bar{m}}^{t,e}} \bbE\left[\left\lVert \nabla \ell(\bfs; \bsth_{r,\bar{m}}^{t,e}) - \nabla L_{\bar{m}}(\bsth_{r,\bar{m}}^{t,e}) \right\rVert^2 \right] \\
        &\le \sigma^2,
    \end{align}
    where the first inequality follows from Jensen's inequality and the second from \eqref{eq:boundness2} in \textbf{(A3)}.
\end{IEEEproof}

\begin{lemma} \label{lem:subs}
    For $\bsth_{r,\bar{m}}^{t,e}$ in \textbf{Algorithm 1},
    \begin{align}
        \bsth_{r,\bar{m}}^{t,e} - \bsth^t = -\lambda \sum_{\mu=1}^{\bar{m}}\sum_{\epsilon=1}^{\calE(\mu)} \nabla \bar{\ell}_{r,\mu}^{t,\epsilon}, \nonumber 
    \end{align}
    where 
    \begin{align}
        \calE(\mu) =
        \begin{cases}
            E&\text{for}~\mu\in[1:\bar{m}-1] \\
            e-1 & \text{for}~\mu=\bar{m}.
        \end{cases} \nonumber 
    \end{align}
\end{lemma}
\begin{IEEEproof}
    \begin{align}
        &\bsth_{r,\bar{m}}^{t,e} \! - \! \bsth^t
        \!=\! \bsth_{r,\bar{m}}^{t,e}  \!+\! \sum_{\mu=1}^{\bar{m}-1}\bsth_{r,\mu}^{t,E\!+\!1} \!-\!\sum_{\mu=1}^{\bar{m}-1}\bsth_{r,\mu}^{t,E\!+\!1} \! - \! 
    \bsth_{r,1}^{t,1}\\
        &= \bsth_{r,\bar{m}}^{t,e} + \sum_{\mu=1}^{\bar{m}-1}\bsth_{r,\mu}^{t,E+1} - \sum_{\mu=1}^{\bar{m}-1}\bsth_{r,\mu+1}^{t,1} - \bsth_{r,1}^{t,1}\\
        &= \bsth_{r,\bar{m}}^{t,e} - \bsth_{r,\bar{m}}^{t,1}+ \sum_{\mu=1}^{\bar{m}-1}\bsth_{r,\mu}^{t,E+1} - \sum_{\mu=1}^{\bar{m}-1}\bsth_{r,\mu}^{t,1}\\ 
        &=\sum_{\mu=1}^{\bar{m}} (\bsth_{r,\mu}^{t,\calE(\mu)+1} - \bsth_{r,\mu}^{t,1})\\
        &=\sum_{\mu=1}^{\bar{m}} \bigg(\bsth_{r,\mu}^{t,\calE(\mu)+1} \!+\!\sum_{\epsilon=2}^{\calE(\mu)} \bsth_{r,\mu}^{t,\epsilon}\! - \!\sum_{\epsilon=2}^{\calE(\mu)} \bsth_{r,\mu}^{t,\epsilon} \!- \! \bsth_{r,\mu}^{t,1}\bigg)\\
        &=\sum_{\mu=1}^{\bar{m}} \sum_{\epsilon=1}^{\calE(\mu)} (\bsth_{r,\mu}^{t,\epsilon+1} -  \bsth_{r,\mu}^{t,\epsilon} )\\
        & = -\lambda \sum_{\mu=1}^{\bar{m}}\sum_{\epsilon=1}^{\calE(\mu)} \nabla \bar{\ell}_{r,\mu}^{t,\epsilon}.
    \end{align}
\end{IEEEproof}

\begin{lemma} \label{lem:subs2}
    For $\bsth_{r,\bar{m}}^{t,e}$ in \textbf{Algorithm 1}, 
    \begin{align}
        &\bbE[\lVert \bsth_{r,\bar{m}}^{t,e} - \bsth^t \rVert^2] \nonumber \\
        &\le 4\lambda^2\zeta_{\bar{m}}^e\sigma^2 + 4\lambda^2 \kappa^2 \zeta_{\bar{m}}^e \sum_{\mu=1}^{\bar{M}}\sum_{\epsilon=1}^{E}\bbE[\lVert 
\bsth_{r,\mu}^{t,\epsilon} - \bsth^t \rVert^2] \nonumber \\
&+4\lambda^2 (\zeta_{\bar{m}}^{e})^2 M \delta^2 + 4\lambda^2 (\zeta_{\bar{m}}^{e})^2 \lVert \nabla L(\bsth^t) \rVert^2,
    \end{align}
    where $\zeta_{\bar{m}}^{e} = E(\bar{m}-1)+e-1$.
\end{lemma}
\begin{IEEEproof}
    By using \textbf{Lemma \ref{lem:subs}}, we can obtain \eqref{lem:pf01}-\eqref{lem:pf02}, where the inequality to \eqref{lem:pf02} is from Cauchy-Schwarz inequality.

\setcounter{equation}{31}

    Firstly, (L.1) in \eqref{lem:pf02} is bounded by
    \begin{align}
        \text{(L.1)}&= \sum_{\mu=1}^{\bar{m}}\sum_{\epsilon=1}^{\calE(\mu)} \bbE[\lVert \nabla \bar{\ell}_{r,\mu}^{t,\epsilon} - \nabla L_{\mu}(\bsth_{r,\mu}^{t,\epsilon}) \rVert^2] \labelrel\le{lem01} \zeta_{\bar{m}}^{e} \sigma^2,  \label{lem:pf03}
    \end{align}
    where inequality \eqref{lem01} holds due to \textbf{Lemma \ref{lem:bound_Loss2}}.  Secondly, (L.2) is bounded by
    \begin{align}
        \text{(L.2)} &\labelrel\le{lem02} \zeta_{\bar{m}}^{e} \sum_{\mu=1}^{\bar{m}}\sum_{\epsilon=1}^{\calE(\mu)} \bbE[\lVert \nabla L_{\mu}(\bsth_{r,\mu}^{t,\epsilon}) - \nabla L_{\mu}(\bsth^t) \rVert^2] \nonumber \\
        &\labelrel\le{lem03} \kappa^2 \zeta_{\bar{m}}^{e} \sum_{\mu=1}^{\bar{m}}\sum_{\epsilon=1}^{\calE(\mu)} \bbE[\lVert \bsth_{r,\mu}^{t,\epsilon} - \bsth^t\rVert^2] \nonumber \\
        &\le \kappa^2 \zeta_{\bar{m}}^{e} \sum_{\mu=1}^{\bar{M}}\sum_{\epsilon=1}^{E} \bbE[\lVert \bsth_{r,\mu}^{t,\epsilon} - \bsth^t\rVert^2], \label{lem:pf04}
    \end{align}
    where inequalities \eqref{lem02} and \eqref{lem03} hold due to Cauchy-Schwarz inequality and \eqref{eq:smooth} in \textbf{Assumption 1}, respectively. Thirdly, (L.3) is bounded by
    \begin{align}
        \text{(L.3)} &\labelrel\le{lem04} \zeta_{\bar{m}}^{e} \sum_{\mu=1}^{\bar{m}}\sum_{\epsilon=1}^{\calE(\mu)} \bbE[\lVert \nabla L_{\mu}(\bsth^t) - \nabla L(\bsth^t) \rVert^2] \\
        &\labelrel\le{lem05} (\zeta_{\bar{m}}^{e})^2 M \delta^2, \label{lem:pf05}
    \end{align}
    where inequalities \eqref{lem04} and \eqref{lem05} hold due to Cauchy-Schwarz inequality and \textbf{Lemma \ref{lem:bound_Loss1}}, respectively. Lastly, (L.4) can be rewritten as
    \begin{align}
        \text{(L.4)}=(\zeta_{\bar{m}}^{e})^2 \lVert \nabla L(\bsth^t) \rVert^2. \label{lem:pf06}
    \end{align}
    By substituting \eqref{lem:pf03}-\eqref{lem:pf06} into \eqref{lem:pf02}, we obtain the desired result.
\end{IEEEproof}

\section{Proof of Lemma \ref{lem:main}}
Firstly, at each epoch $t$, there is at least one cluster $\calQ_r^t$ including no malicious client due to the pigeonhole principle. 

From now on, the expectation is conditioned on $\bsth^t$, and we omit the this condition unless otherwise specified for ease of notation. 
Note that 
\begin{align}
    \bbE[L(\bsth^{t+1}) - L(\bsth^{t})] 
    &=\bbE[L(\bsth_{\hat{r}}^{t+1}) - L(\bsth^{t})] \nonumber \\
    &\labelrel={a01}\bbE[\bar{\ell}_{\hat{r}}^t - L(\bsth^{t})] \nonumber \\ 
    &\labelrel\le{a02} \bbE [\bar{\ell}_{r}^t - L(\bsth^{t})] \nonumber \\ &\labelrel\le{a03}\bbE[L(\bsth_r^{t+1}) - L(\bsth^t)]\label{app:pf01}
\end{align}
where equalities \eqref{a01} and \eqref{a03} hold due to \textbf{Lemma \ref{lem:loss_exp}}, and inequality \eqref{a02} hold 
since $\bar{l}_r^t$ representing the loss of cluster $\calQ_r^t$ is greater than $\bar{l}_{\hat{r}}^{t}$ denoting one of $\calQ_{\hat{r}}^t$. Based on \eqref{app:pf01}, we can obtain  
\begin{align}
    &\bbE[L(\bsth^{t+1}) - L(\bsth^{t})] 
    \le \bbE[L(\bsth_r^{t+1}) - L(\bsth^t)] \nonumber \\
    &\labelrel\le{a04} \underbrace{\bbE[\nabla L(\bsth^{t})^{\mathsf{T}}(\bsth_{r}^{t+1} - \bsth^t)]}_{\triangleq \text{(A.1)}} + \frac{\kappa}{2}\cdot\underbrace{\bbE[\lVert \bsth_r^{t+1} -\bsth^t\rVert^2]}_{\triangleq \text{(A.2)}}, \label{app:pf02} 
\end{align}
where \eqref{a04} holds due to $\kappa$-smoothness of $L_m$'s described in \eqref{eq:smooth2} in \textbf{Assumption 1}. (A.1) can be rewritten as
\begin{align}
    &\text{(A.1)} \labelrel={a05} -\lambda \bbE\bigg[\nabla L(\bsth^t)^{\mathsf{T}}\sum_{\bar{m}=1}^{\bar{M}}\sum_{e=1}^{E}\nabla \bar{\ell}_{r,\bar{m}}^{t,e}\bigg] \nonumber \\
    &=-\lambda\bar{M}\sum_{e=1}^{E} \bbE\bigg[ \nabla L(\bsth^t)^\mathsf{T} \bigg( \frac{1}{\bar{M}}\sum_{\bar{m}=1}^{\bar{M}} \nabla\bar{\ell}_{r,\bar{m}}^{t,e}\bigg)\bigg] \nonumber  \\
    &\labelrel={a06} -\lambda\bar{M}\sum_{e=1}^{E} \nabla L(\bsth^t)^\mathsf{T} \bbE\bigg[\frac{1}{\bar{M}}\sum_{\bar{m}=1}^{\bar{M}} \nabla L_{\bar{m}}(\bsth_{r,\bar{m}}^{t,e})\bigg] \nonumber  \\
    &= -\frac{\lambda\bar{M}}{2} \sum_{e=1}^{E} \bigg(\lVert\nabla L(\bsth^t)\rVert^2 \nonumber \\
    &\qquad+ \bigg\lVert\bbE\bigg[  \frac{1}{\bar{M}}\sum_{\bar{m}=1}^{\bar{M}} \nabla L_{\bar{m}}(\bsth_{r,\bar{m}}^{t,e})\bigg]\bigg\rVert^2  \nonumber \\
    &\qquad- \bigg\lVert \bbE\bigg[\frac{1}{\bar{M}}\sum_{\bar{m}=1}^{\bar{M}} \nabla L_{\bar{m}}(\bsth_{r,\bar{m}}^{t,e})\bigg] - \nabla L(\bsth^t)  \bigg\rVert^2\bigg) \nonumber \\
    &=-\frac{\lambda\bar{M}}{2} \sum_{e=1}^{E} \bigg(\lVert\nabla L(\bsth^t)\rVert^2 \nonumber \\
    &\qquad\qquad\qquad+ \bigg\lVert\bbE\bigg[  \frac{1}{\bar{M}}\sum_{\bar{m}=1}^{\bar{M}} \nabla L_{\bar{m}}(\bsth_{r,\bar{m}}^{t,e})\bigg]\bigg\rVert^2\bigg) \nonumber \\ 
    &+ \frac{\lambda \bar{M}}{2} \sum_{e=1}^{E} \underbrace{\bigg\lVert\bbE\bigg[ \frac{1}{\bar{M}}\sum_{\bar{m}=1}^{\bar{M}} \nabla L_{\bar{m}}(\bsth_{r,\bar{m}}^{t,e})\bigg] - \nabla L(\bsth^t)  \bigg\rVert^2}_{\triangleq \text{(B)}}, \label{app:pf03}
\end{align}
where \eqref{a05} holds due to \textbf{Lemma \ref{lem:subs}} for $\bar{m}=\bar{M}$ and $e=E+1$, and \eqref{a06} comes from \eqref{eq:sgd_property} in \textbf{Assumption 2}.

Again, (B) can be bounded by
\begin{align}
    &\text{(B)} = \bigg\lVert\bbE\bigg[ \frac{1}{\bar{M}}\sum_{\bar{m}=1}^{\bar{M}} \big(\nabla L_{\bar{m}}(\bsth_{r,\bar{m}}^{t,e})- \nabla L(\bsth^t)\big)\bigg]   \bigg\rVert^2 \nonumber \\
    &=\bigg\lVert\bbE\bigg[ \frac{1}{\bar{M}}\sum_{\bar{m}=1}^{\bar{M}} \big(\nabla L_{\bar{m}}(\bsth_{r,\bar{m}}^{t,e}) - \nabla L_{\bar{m}}(\bsth^t) \nonumber \\
    &\qquad \qquad \qquad \qquad + \nabla L_{\bar{m}}(\bsth^t)- \nabla L(\bsth^t)\big)\bigg]   \bigg\rVert^2 
    \nonumber \\
    &= \bigg\lVert\bbE\bigg[ \frac{1}{\bar{M}}\sum_{\bar{m}=1}^{\bar{M}} \big(\nabla L_{\bar{m}}(\bsth_{r,\bar{m}}^{t,e})- \nabla L_{\bar{m}}(\bsth^t)\big)\bigg]   \bigg\rVert^2 \nonumber \\
    &\labelrel\le{a07} \frac{1}{\bar{M}}\sum_{\bar{m}=1}^{\bar{M}}\bbE[\lVert  \nabla L_{\bar{m}}(\bsth_{r,\bar{m}}^{t,e}) - \nabla L_{\bar{m}}(\bsth^t)\rVert^2] \nonumber \\
    &\labelrel\le{a09} \frac{\kappa^2}{\bar{M}} \sum_{\bar{m}=1}^{\bar{M}} \bbE[\lVert \bsth_{r,\bar{m}}^{t,e} - \bsth^t \rVert^2] \label{app:pf04}
\end{align}
where inequalities \eqref{a07} and \eqref{a09} hold due to Jensen's inequality and $\kappa$-smoothness of $L_{\bar{m}}$'s described in \eqref{eq:smooth} in \textbf{Assumption 1}, respectively. Go back to (A.2), we can obtain
\begin{align}
    &\text{(A.2)}= \lambda^2 \bbE\bigg[\bigg\lVert \sum_{\bar{m}=1}^{\bar{M}}\sum_{e=1}^{E} \big( \nabla \bar{l}_{r,\bar{m}}^{t,e} - \nabla L_{\bar{m}}(\bsth_{r,\bar{m}}^{t,e}) \nonumber \\
    &\qquad\qquad\qquad\qquad\qquad\qquad\qquad+ \nabla L_{\bar{m}}(\bsth_{r,\bar{m}}^{t,e}) \big) \bigg\rVert^2\bigg] \nonumber \\
    &\labelrel\le{a10} 2\lambda^2 \bbE\bigg[\bigg\lVert \sum_{\bar{m}=1}^{\bar{M}}\sum_{e=1}^{E} \big( \nabla \bar{l}_{r,\bar{m}}^{t,e} - \nabla L_{\bar{m}}(\bsth_{r,\bar{m}}^{t,e})\big) \bigg\rVert^2\bigg] \nonumber \\
    &\qquad\qquad\qquad\qquad+2\lambda^2 \bbE\bigg[\bigg\lVert \sum_{\bar{m}=1}^{\bar{M}}\sum_{e=1}^{E}  \nabla L_{\bar{m}}(\bsth_{r,\bar{m}}^{t,e}) \bigg\rVert^2\bigg] \nonumber \\
    &\labelrel\le{a11} 2\lambda^2 \bar{M}E \sum_{\bar{m}=1}^{\bar{M}}\sum_{e=1}^{E} \bbE[\lVert \nabla \bar{l}_{r,\bar{m}}^{t,e} - \nabla L_{\bar{m}}(\bsth_{r,\bar{m}}^{t,e}) \rVert^2] \nonumber \\
    &\qquad\qquad+2\lambda^2 \bar{M}^2 E \sum_{e=1}^{E} \bbE\bigg[\bigg\lVert \frac{1}{\bar{M}}\sum_{\bar{m}=1}^{\bar{M}}\nabla L_{\bar{m}}(\bsth_{r,\bar{m}}^{t,e}) \bigg\rVert^2\bigg] \nonumber \\
    &\labelrel\le{a12} 2\lambda^2 \bar{M}^2 E \nonumber \\
    &~~\times \bigg( E\sigma^2 +  \sum_{e=1}^{E} \underbrace{\bbE\bigg[\bigg\lVert \frac{1}{\bar{M}}\sum_{\bar{m}=1}^{\bar{M}}\nabla L_{\bar{m}}(\bsth_{r,\bar{m}}^{t,e}) \bigg\rVert^2\bigg]}_{\triangleq\text{(C)}} \bigg) \label{app:pf05}
\end{align}
where inequalities \eqref{a10}, \eqref{a11} and \eqref{a12} hold due to the fact that $\lVert 
\bfa+\bfb \rVert^2 \le 2\lVert\bfa\rVert^2 + 2\lVert\bfb\rVert^2$ for $\bfa,\bfb\in\bbR^d$, Jensen's inequality and \textbf{Lemma \ref{lem:bound_Loss2}}, respectively. Then, (C) is bounded by
\begin{align}
    &\text{(C)}\labelrel\le{a12_1}\frac{1}{\bar{M}} \sum_{\bar{m}=1}^{\bar{M}} \bbE[ \lVert  \nabla L_{\bar{m}} (\bsth_{r,\bar{m}}^{t,e}) - \nabla L_{\bar{m}}(\bsth^t)  \nonumber \\
    &\qquad\qquad\qquad+ \nabla L_{\bar{m}}(\bsth^t)- \nabla L(\bsth^t) + \nabla L(\bsth^t)   \rVert^2 ] \nonumber \\ 
    &\labelrel\le{a12_2} \frac{3}{\bar{M}} \sum_{\bar{m}=1}^{\bar{M}} 
    \bbE[\lVert \nabla L_{\bar{m}} (\bsth_{r,\bar{m}}^{t,e}) - \nabla L_{\bar{m}}(\bsth^t) \rVert^2  \nonumber \\
    &\qquad\qquad+\lVert \nabla L_{\bar{m}}(\bsth^t)- \nabla L(\bsth^t) \rVert^2 + \lVert \nabla L(\bsth^t) \rVert^2] \nonumber \\
    &\labelrel\le{a12_3} \frac{3\kappa^2}{\bar{M}} \sum_{\bar{m}=1}^{\bar{M}}\bbE[\lVert \bsth_{r,\bar{m}}^{t,e} - \bsth^t \rVert^2] + 3M\delta^2 + 3 \lVert \nabla L(\bsth^t) \rVert^2, \label{app:pf05_1}
\end{align}
where inequalities \eqref{a12_1}, \eqref{a12_2} and \eqref{a12_3} hold due to Jensen's inequality, Cauchy-Scharwz inequality and $\kappa$-smoothness of $L_{\bar{m}}$'s described in \eqref{eq:smooth} in \textbf{Assumption 1} and \textbf{Lemma \ref{lem:bound_Loss1}}, respectively.
By substituting \eqref{app:pf04} to \eqref{app:pf03} and \eqref{app:pf05_1} to \eqref{app:pf05}, respectively, then \eqref{app:pf03} and \eqref{app:pf05} to \eqref{app:pf02}, we have
\begin{align}
    &\bbE[L(\bsth^{t+1}) - L(\bsth^t)] \le \nonumber \\ 
    &-\frac{\lambda \bar{M}E}{2}(1 - 6\lambda\kappa\bar{M} E) \lVert \nabla L(\bsth^t) \rVert^2 \nonumber \\
    &+\frac{\lambda \kappa^2}{2}(1 + 6\lambda \kappa \bar{M} E)\sum_{e=1}^{E}\sum_{\bar{m}=1}^{\bar{M}} \bbE[\lVert \bsth_{r,\bar{m}}^{t,e} - \bsth^{t} \rVert^2] \nonumber \\
    & -\frac{\lambda\bar{M}}{2} \sum_{e=1}^{E} \bigg\lVert \bbE\bigg[ \frac{1}{\bar{M}}\sum_{\bar{m}=1}^{\bar{M}} \nabla L_{\bar{m}}(\bsth_{r,\bar{m}}^{t,e}) \bigg] \bigg\rVert^2  \nonumber \\
    &+ \lambda^2\kappa \bar{M}^2E^2\sigma^2 + 3\lambda^2 \kappa M \bar{M}^2E^2\delta^2. \label{app:pf06}
\end{align}
Since $0 <\lambda \le \frac{1}{12\kappa \bar{M}E}$, it implies $1-6\lambda\kappa\bar{M}E \ge \frac{1}{2}$ and $1+6\lambda\kappa\bar{M}E\le \frac{3}{2}$. Therefore, \eqref{app:pf06} is reduced to
\begin{align}
    &\bbE[L(\bsth^{t+1}) - L(\bsth^t)] \le  
    -\frac{\lambda \bar{M}E}{4} \lVert \nabla L(\bsth^t) \rVert^2 \nonumber \\
    &+\frac{3\lambda \kappa^2}{4}\sum_{e=1}^{E}\sum_{\bar{m}=1}^{\bar{M}} \bbE[\lVert \bsth_{r,\bar{m}}^{t,e} - \bsth^{t} \rVert^2] \nonumber \\
    &+ \lambda^2\kappa \bar{M}^2E^2\sigma^2 + 3\lambda^2 \kappa M \bar{M}^2E^2\delta^2. \label{app:pf07}
\end{align}
From the second term of the right-hand side in \eqref{app:pf06}, we can obtain the following result.
\begin{align}
    &\sum_{\bar{m}=1}^{\bar{M}}\sum_{e=1}^{E}\bbE[\lVert \bsth_{r,\bar{m}}^{t,e} - \bsth^t \rVert^2]
    \labelrel\le{a13} 4\lambda^2\sigma^2\sum_{\bar{m}=1}^{\bar{M}}\sum_{e=1}^{E}\zeta_{\bar{m}}^{e} \nonumber \\
    &+ 4\lambda^2\kappa^2 \sum_{\bar{m}=1}^{\bar{M}}\sum_{e=1}^{E}  \zeta_{\bar{m}}^{e} \sum_{\mu=1}^{\bar{M}}\sum_{\epsilon=1}^{E} \bbE[\lVert \bsth_{r,\mu}^{t,\epsilon} - \bsth^t \rVert^2] \nonumber \\
    &+4\lambda^2 M\delta^2 \sum_{\bar{m}=1}^{\bar{M}}\sum_{e=1}^{E}(\zeta_{\bar{m}}^{e})^2 \nonumber \\
    &+ 4\lambda^2 \lVert L(\bsth^t) \rVert^2 \sum_{\bar{m}=1}^{\bar{M}}\sum_{e=1}^{E}(\zeta_{\bar{m}}^{e})^2 \nonumber \\
    &\labelrel\le{a14} 4\lambda^2 \bar{M}^2 E^2 \sigma^2 + 4\lambda^2\kappa^2 \bar{M}^2 E^2 \nonumber \\
    &\qquad\times \sum_{\mu=1}^{\bar{M}}\sum_{\epsilon=1}^{E} \bbE[\lVert \bsth_{r,\mu}^{t,\epsilon} - \bsth^t \rVert^2] \nonumber \\
    &\qquad +4\lambda^2 M\bar{M}^3 E^3 \delta^2 + 4\lambda^2 \bar{M}^3 E^3 \lVert 
 \nabla L(\bsth^t) \rVert^2,  \label{app:pf08}
\end{align}
where inequalities \eqref{a13} and \eqref{a14} hold due to \textbf{Lemma \ref{lem:subs2}} and 
\begin{align}
    &\sum_{\bar{m}=1}^{\bar{M}} \sum_{e=1}^{E} \zeta_{\bar{m}}^{e} \le \bar{M}^2 E^2, \nonumber \\
    &\sum_{\bar{m}=1}^{\bar{M}} \sum_{e=1}^{E} (\zeta_{\bar{m}}^{e})^2 \le \bar{M}^3 E^3, \nonumber 
\end{align}
respectively.
Then, by reorganizing the above inequality \eqref{app:pf08}, we can obtain the following relation.
\begin{align}
    &(1-4\lambda^2 \kappa^2 \bar{M}^2 E^2 ) \sum_{\bar{m}=1}^{\bar{M}} \sum_{e=1}^{E} \bbE[\lVert \bsth_{r,\bar{m}}^{t,e} - \bsth^t \rVert^2] \nonumber \\
    &\qquad\qquad\le 4\lambda^2\bar{M}^2 E^2 \sigma^2 + 4\lambda^2 M \bar{M}^3 E^3 \delta^2 \nonumber \\
    &\qquad\qquad\qquad+ 4\lambda^2 \bar{M}^3 E^3 \lVert\nabla L(\bsth^t)\rVert^2.  \nonumber 
\end{align}
Since $\lambda\le\frac{1}{12\kappa\bar{M}E}\le \frac{1}{4\kappa \bar{M} E}$, it implies $1-4\lambda^2 \kappa^2 \bar{M}^2 E^2 \ge \frac{3}{4}\ge\frac{1}{2}$,
and we can obtain
\begin{align}
    &\sum_{\bar{m}=1}^{\bar{M}}\sum_{e=1}^{E} \bbE[\lVert \bsth_{r,\bar{m}}^{t,e} - \bsth^t\rVert^2] \le  8\lambda^2 \bar{M}^2 E^2 \sigma^2 \nonumber \\
    &\qquad+ 8\lambda^2 M\bar{M}^3 E^3 \delta^2 + 8\lambda^2 \bar{M}^3 E^3 \lVert\nabla L(\bsth^t)\rVert^2. \label{app:pf09} 
\end{align}
Again, by substituting \eqref{app:pf09} to \eqref{app:pf07},
\begin{align}
    &\bbE[L(\bsth^{t+1}) - L(\bsth^t)] \nonumber \\
    &\le  
    -\frac{\lambda \bar{M}E}{4} (1 - 24\lambda^2 \kappa^2 \bar{M}^2 E^2) \lVert \nabla L(\bsth^t) \rVert^2  \nonumber \\
    &+6\lambda^3\kappa^2 M\bar{M}^3 E^3 \delta^2 + 6\lambda^3 \kappa^2 \bar{M}^2 E^2 \sigma^2 \nonumber \\ 
    &+ \lambda^2\kappa \bar{M}^2E^2\sigma^2 + 3\lambda^2 \kappa M \bar{M}^2E^2\delta^2. \label{app:pf10}
\end{align}
Since $\lambda \le \frac{1}{12\kappa\bar{M}E} \le \frac{1}{6\kappa \bar{M} E}$ implies $1 - 24\lambda^2 \kappa^2 \bar{M}^2 E^2 \ge \frac{1}{3}$, we can obtain the following desired result.
\begin{align}
    &\bbE[L(\bsth^{t+1})] - L(\bsth^t) \le -\frac{\lambda \bar{M}E}{12} \lVert  L(\bsth^t)\rVert^2 \nonumber \\
    &\qquad\qquad+6\lambda^3\kappa^2 M\bar{M}^3 E^3 \delta^2 + 6\lambda^3 \kappa^2 \bar{M}^2 E^2 \sigma^2 \nonumber \\ 
    &\qquad\qquad+ \lambda^2\kappa \bar{M}^2E^2\sigma^2 + 3\lambda^2 \kappa M \bar{M}^2E^2\delta^2.
\end{align}

\section{Proof of Theorem \ref{thm:main}}
Based on \textbf{Lemma \ref{lem:main}}, by summing up from $t=1$ to $T$, we can obtain
\begin{align}
    &\bbE[L(\bsth^{T+1})] - L(\bsth^1) 
    \le -\frac{\lambda \bar{M} E}{12} \sum_{t=1}^{T} \bbE[\lVert L(\bsth^t) \rVert^2] \nonumber \\
    &\qquad+6\lambda^3 T\kappa^2 M\bar{M}^3 E^3 \delta^2 + 6\lambda^3 T \kappa^2 \bar{M}^2 E^2 \sigma^2 \nonumber \\ 
    &\qquad+ \lambda^2  T\kappa \bar{M}^2E^2\sigma^2 + 3\lambda^2 T \kappa M \bar{M}^2E^2\delta^2. \label{app2:01}
\end{align}
By reorganizing \eqref{app2:01} and dividing both sides by $T$, 
\begin{align}
    &\frac{1}{T} \sum_{t=1}^{T} \bbE[\lVert L(\bsth^t) \rVert^2] 
    \le -\frac{12}{\lambda T\bar{M}E} \big(\bbE[L(\bsth^{T+1})] - L(\bsth^1)\big) \nonumber \\
    &\qquad+72\lambda^2 \kappa^2 M \bar{M}^2 E^2 \delta^2 + 72 \lambda^2 \kappa^2 \bar{M} E \sigma^2 \nonumber \\
    &\qquad+12\lambda \kappa \bar{M} E \sigma^2 + 36\lambda \kappa M \bar{M} E \delta^2. \label{app2:02}
\end{align}
When $\lambda = \frac{1}{12\kappa\bar{M}E\sqrt{T}}$, \eqref{app2:02} is reduced to the desired result as follows:
\begin{align}
    &\frac{1}{T} \sum_{t=1}^{T} \bbE[\lVert L(\bsth^t) \rVert^2] 
    \le -\frac{144\kappa}{\sqrt{T}} \big(\bbE[L(\bsth^{T+1})] - L(\bsth^1)\big) \nonumber \\
    &\ +\!\frac{M  \delta^2}{2T} \! +\!\frac{R\sigma^2}{2MET}\! +\!\frac{\sigma^2}{\sqrt{T}} \!+\! \frac{3M\delta^2}{\sqrt{T}}\!=\! \calO\bigg(\frac{1}{\sqrt{T}}\bigg).
\end{align}


%




\ifCLASSOPTIONcaptionsoff
  \newpage
\fi



%

\bibliographystyle{IEEEtran}
\bibliography{IEEEabrv,CSL}

\end{document}